\documentclass[letterpaper, 10 pt, conference]{ieeeconf}  

\IEEEoverridecommandlockouts                              
\usepackage{scalerel}
\usepackage{tikz}
\usepackage{amsmath}
\usetikzlibrary{svg.path}
\usepackage{colortbl}
\usepackage{paralist}
\usepackage{framed}

\definecolor{orcidlogocol}{HTML}{A6CE39}
\tikzset{
  orcidlogo/.pic={
    \fill[orcidlogocol] svg{M256,128c0,70.7-57.3,128-128,128C57.3,256,0,198.7,0,128C0,57.3,57.3,0,128,0C198.7,0,256,57.3,256,128z};
    \fill[white] svg{M86.3,186.2H70.9V79.1h15.4v48.4V186.2z}
                 svg{M108.9,79.1h41.6c39.6,0,57,28.3,57,53.6c0,27.5-21.5,53.6-56.8,53.6h-41.8V79.1z M124.3,172.4h24.5c34.9,0,42.9-26.5,42.9-39.7c0-21.5-13.7-39.7-43.7-39.7h-23.7V172.4z}
                 svg{M88.7,56.8c0,5.5-4.5,10.1-10.1,10.1c-5.6,0-10.1-4.6-10.1-10.1c0-5.6,4.5-10.1,10.1-10.1C84.2,46.7,88.7,51.3,88.7,56.8z};
  }
}

\newcommand\orcidicon[1]{\href{https://orcid.org/#1}{\mbox{\scalerel*{
\begin{tikzpicture}[yscale=-1,transform shape]
\pic{orcidlogo};
\end{tikzpicture}
}{|}}}}

\usepackage{hyperref}

\title{\LARGE \bf
  Transparency in Multi-Human Multi-Robot Interaction} 
\author{Jayam Patel$^{1}$\orcidicon{0000-0002-0687-4169}, Tyagaraja Ramaswamy$^{1}$, Zhi Li$^{1}$ and Carlo Pinciroli$^{1}$\orcidicon{0000-0002-2155-0445} 
\thanks{$^{1}$ Department of Robotics Engineering, Worcester Polytechnic Institute, MA, USA.
Email: \texttt{\{jupatel}, \texttt{tramaswamy}, \texttt{zli11}, \texttt{cpinciroli\}@wpi.edu}}}

\usepackage{graphicx}
\usepackage{subcaption}
\usepackage{multirow}
\usepackage{units}
\usepackage{hyperref}
\begin{document}

\maketitle
\thispagestyle{empty}
\pagestyle{empty}


\begin{abstract}
Transparency is a key factor in the performance of human-robot interaction. A transparent interface allows operators to be aware of the state of a robot and to assess the progress of the tasks at hand. When multi-robot systems are involved, transparency is a greater challenge, due to the larger number of variables affecting the behavior of the robots as a whole. Existing work studies transparency with single operators and multiple robots. Studies on transparency that focus on \textit{multiple} operators interacting with a multi-robot systems are limited. This paper fills this gap by presenting a novel human-swarm interface for multiple operators. Through this interface, we study which graphical elements are contributing to multi-operator transparency by comparing four ``transparency modes'':
\begin{inparaenum}[\it (i)]
\item no transparency (no operator receives information from the robots),
\item central transparency (the operators receive information only relevant to their personal task),
\item peripheral transparency (the operators share information on each others' tasks), and
\item mixed transparency (both central and peripheral).
\end{inparaenum}
We report the results in terms of awareness, trust, and workload from a user study involving 18 participants engaged in a complex multi-robot task.
\end{abstract}

\section{Introduction}


Human-robot teams are often envisioned in complex scenarios~\cite{Brambilla2013}, including humanitarian missions~\cite{murphy2014disaster,hamins2015research}, interplanetary exploration~\cite{goldsmith1999book}, ecosystem restoration~\cite{denkenberger2007comparison, buters2019methodological}, mining~\cite{rubio2012mining}, bridge inspection~\cite{oh2009bridge} and surgery~\cite{sirouspour2005multi}. The success of these missions depends on effective and efficient team interaction. One crucial requirement to make this vision a reality is making the multi-robot system more transparent~\cite{roundtree_transparency:_2019}, i.e., legible and interpretable, for the human operators.

Transparency is a key property of any human-machine interface. Transparent interfaces offer high usability and foster increased situational awareness \cite{roundtree_transparency:_2019, bhaskara_agent_2020, chakraborti_explicability?_nodate, chen_situation_2018, tulli_eects_nodate}. Transparent interfaces limit or remove ambiguity, improve trust, and enhance decision-making \cite{kalpagam_ganesan_better_2018, de_paolis_debugging_2019}. Lyon’s models of transparency~\cite{lyons2013being} and the situational awareness-based transparency (SAT)~\cite{chen_situation_2014} model provide guidelines for an effective interaction between an operator and a machine.

However, these models are designed and tested with a single operator in mind. The problem of designing a transparent interface intensifies when there are multiple human operators, or `the machine' is, in fact, a multi-robot system. This is because the heterogeneous nature and sheer number of combined interactions among operators and robots affect the behavior and the performance of the entire system in non-trivial ways. As an example, imagine an automated warehouse in which hundreds of robots navigate and transport heavy objects. The robots might drop objects, experience hardware failure, or transport an incorrect object. To resolve these issues, the operators must collaborate and resolve the issues using information from robots and other operators.

In this paper, we study the information that each human operator must process and use, which affects cognitive load~\cite{miller1956magical, lewis2010choosing}. To decrease cognitive load, a possible approach is to limit the amount and the type of information presented to the operator. However, this creates a trade-off with transparency, which intuitively suggests more information would be better.

We explore the design space of graphical user interfaces for human-robot interaction, focusing on the \textit{multi}-human \textit{multi}-robot scenario, which has, so far, received limited attention. We consider four types of interfaces, each presenting different amounts and types of information, and each corresponding to a specific `transparency mode'.

\begin{figure}[t]
    \centering
    \includegraphics[width=0.7\linewidth]{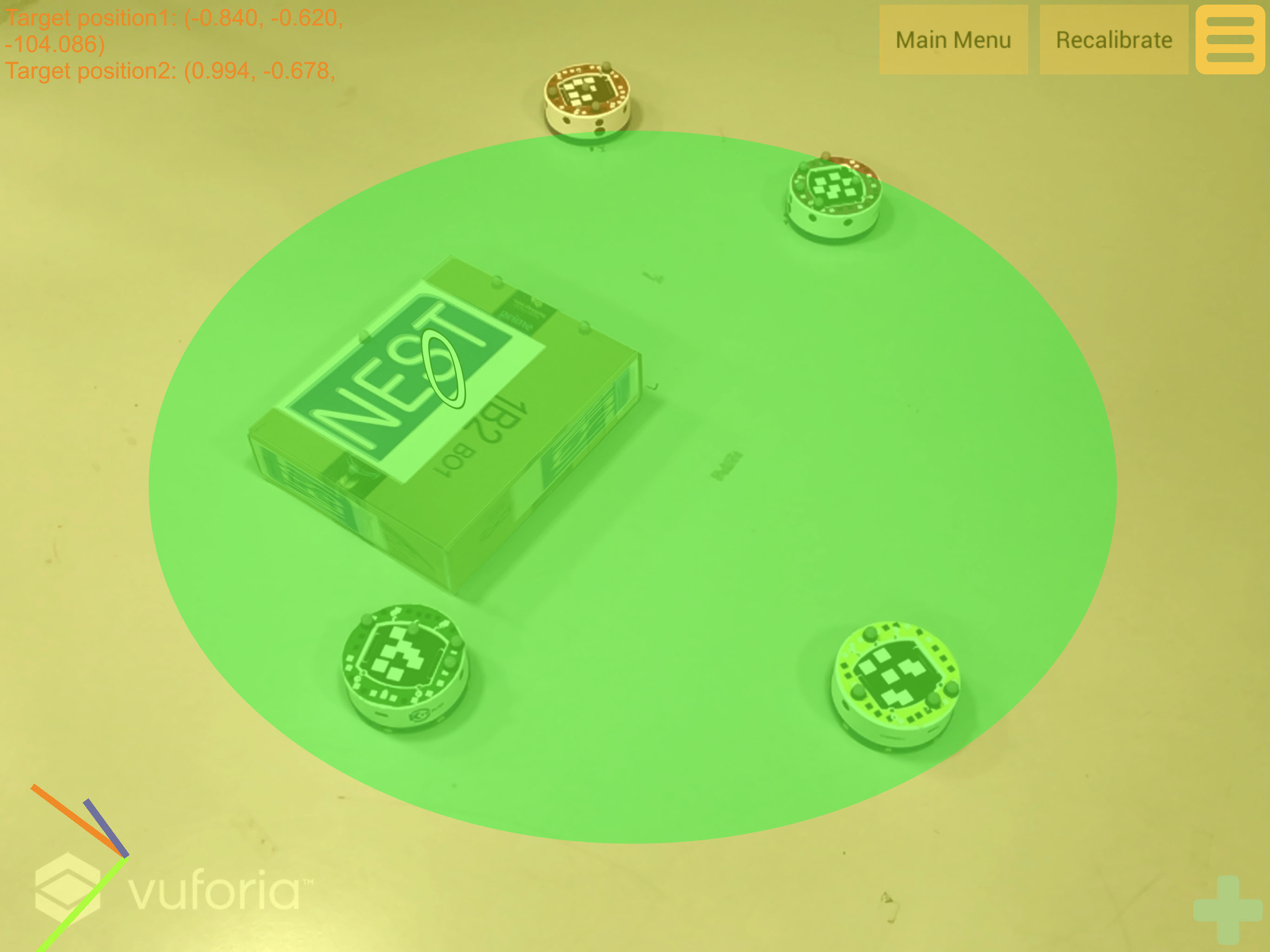}
    \caption{Central and peripheral regions of the field of view. The green region indicates the central field of view. The yellow region indicates the peripheral field of view.}
    \label{fig-transparency:fov}
\end{figure}
To characterize these modes, we study the effect of the field of view (FoV) an interface offers to the operators. We define the FoV as the observable area an operator can see through the interface camera. As shown in Fig.~\ref{fig-transparency:fov}, we categorize the FoV into two regions, based on the distance from the center of the screen: central and peripheral. The \textit{central} FoV is the region closest to the center; the \textit{peripheral} FoV is the remaining region. Using this categorization, we produced four types of interfaces, each differing in the way information is displayed:
\begin{itemize}
\item \textbf{No Transparency (NT):} no information is available to the operator, used as a baseline for comparison;
\item \textbf{Central Transparency (CT):} information is available at the center of the FoV and displayed directly on the robots;
\item \textbf{Peripheral Transparency (PT):} information is shown at the boundaries of the FoV as dedicated widgets;
\item \textbf{Mixed Transparency (MT):} a combination of central and peripheral transparency.
\end{itemize}

We investigate the effects of the transparency modes on operator performance, awareness, task load, and trust in the system. This paper offers two main contributions:
\begin{enumerate}
\item A novel augmented-reality-based interface for multi-human multi-robot interaction with the mentioned transparency modes. This interface is an improvement of our mixed-granularity control interface~\cite{patel2019, patel2019improving} for single operators;
\item A study, which to the best of our knowledge is the first, of the effects of transparency in multi-human multi-robot interaction. Our user study involved 18 participants in teams of 2, each team controlling 9 robots in an object transport scenario.
\end{enumerate}

The paper is organized as follows. In Sec.~\ref{sec-transparency:background}, we discuss related work on transparency. In Sec.~\ref{sec-transparency:framework}, we present our system and its design. In Sec.~\ref{sec-transparency:userstudy}, we report our user study procedures and results followed by analysis in Sec.~\ref{sec-transparency:discussion} and summarize the paper in Sec.~\ref{sec-transparency:conclusion}.
 
\section{Background}
\label{sec-transparency:background}
Transparency is an important research topic in human-machine and human-robot interaction~\cite{schmorrow_proposed_2016,felzmann_robots_2019}. Transparency affects usability~\cite{chien_influence_2019, zhu_enhancing_2020, panganiban_transparency_2019}, performance~\cite{chen_increasing_2015, lakhmani_exploring_2019}, trust~\cite{matthews_individual_2019, guznov_robot_2019} and explainability~\cite{daily_world_2003,wright_transparency_2015}. The effect of these factors increases with the type and quantity of information provided to the operator~\cite{wright2017agent, wright_effects_2015}. Coarse information often negatively affects decision time, trust, situational awareness, and performance; in contrast, detailed information typically results in higher cognitive load.

Ghiringhelli \textit{et al.}~\cite{ghiringhelli_interactive_2014} first proposed to graphically represent the actions of the robots using augmented reality for single operators. Chen \textit{et al.}~\cite{chen_situation_2018} and Mercado \textit{et al.}~\cite{mercado_intelligent_2016} tested the impact of transparency on situational awareness, trust, and workload of an operator. Their work is based on simulated point-mass models of the robots, which lack important physical properties of mobile robots and create a `reality gap' between results collected with a simulated environment and the results collected with physical environment~\cite{jakobi1995noise}. 

With multiple operators, a novel problem arises: the need for operators to share robots and their information, to achieve a new form of transparency which we call \textit{operator-level transparency}. To the best of our knowledge, there is no study on this topic in the context of multi-robot systems controlled through augmented reality (AR).

\section{Transparency-based Interaction System} \label{sec-transparency:framework}
\subsection{System Overview}

\begin{figure}[t]
    \centering
    \includegraphics[width=0.7\linewidth]{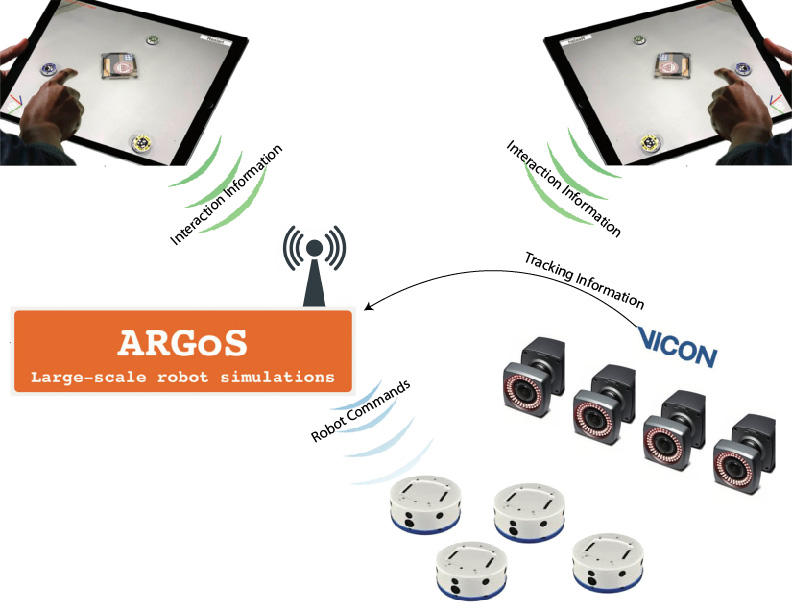}
    \caption{System overview.}
    \label{fig-transparency:system_overview}
\end{figure}

Our system comprises four components (see Fig.~\ref{fig-transparency:system_overview}):
\begin{enumerate}
\item A distributed AR interface implemented as an app for an Apple iPad;
\item A team of robots, pre-programmed with various behaviors to reach a defined point, recognize objects, and perform collective transport;
\item Vicon~\cite{vicon}, a motion capture system that localizes the robots and the movable objects in the environment;
\item ARGoS~\cite{Pinciroli:SI2012}, a multi-robot simulator modified to act as `software glue' between the app, the robots, and the Vicon. We replaced the simulated physics engine shipped with ARGoS with a plug-in that receives positional data from the Vicon motion capture system, and developed new sensor and actuator plug-ins that interface with those on-board the robots. With these plug-ins, ARGoS acts as a middleware functionally similar to the \texttt{roscore} of ROS.
\end{enumerate}

\subsection{User Interface}
\begin{figure}[t]
\centering
\includegraphics[width=0.7\linewidth]{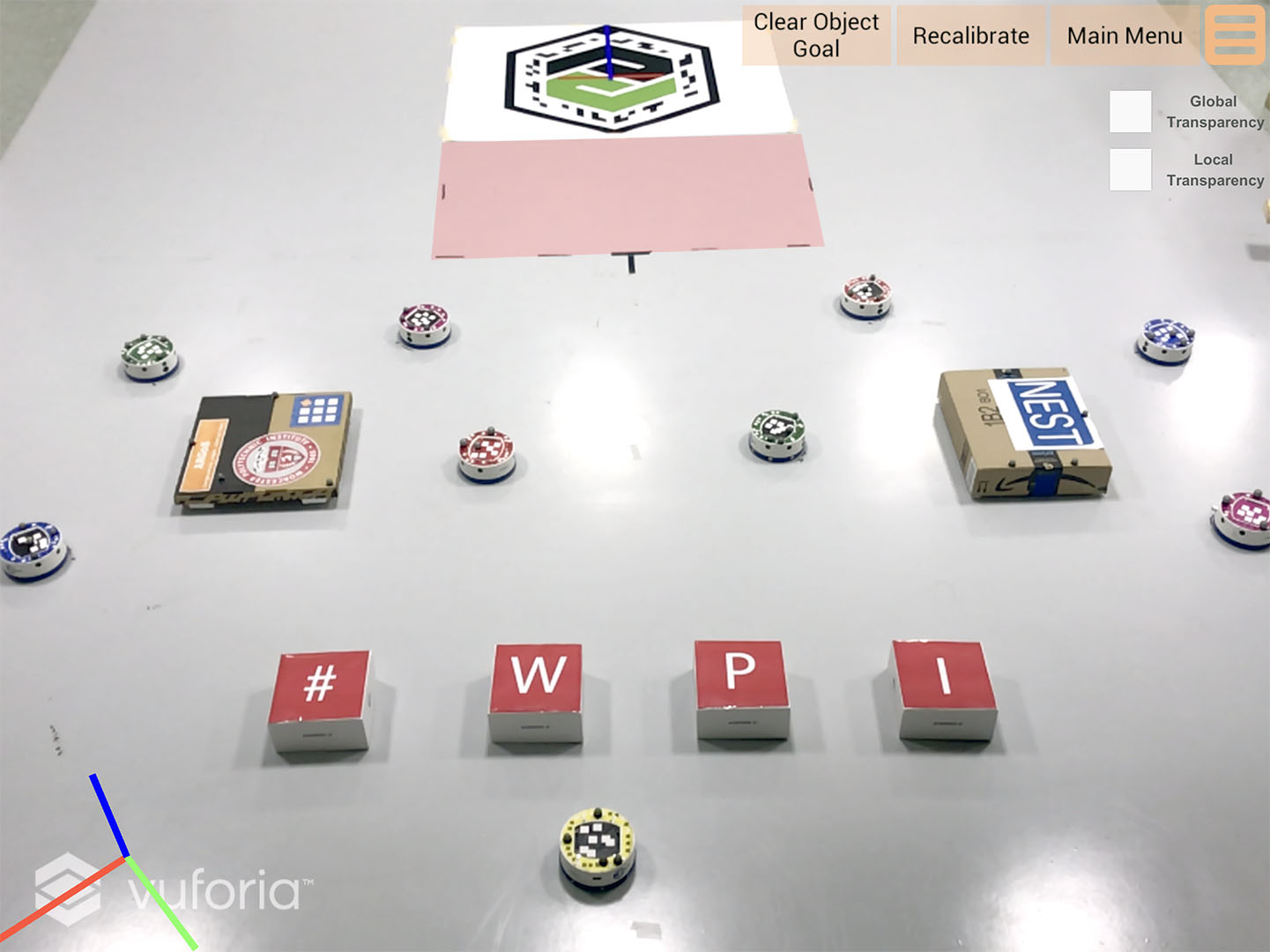}
\caption{Screenshot of our augmented-reality interface running on an iPad. The black arrow indicates the origin marker that corresponds to the coordinate frame of the interface.}
\label{fig-transparency:screenshot}
\end{figure}

Our interface integrates an AR software development kit, Vuforia~\cite{vuforia}, and the Unity~\cite{engine2008unity} game engine. The interface detects robots and movable objects by their fiducial markers. The robots and objects recognized by the interface are overlaid by virtual objects. The operator can manipulate the virtual objects to send commands to the robots. For example, the operator can translate a virtual object with a one-finger swipe and rotate it with a two-finger twist. It is also possible to select a team of robots by drawing a closed path with a continuous one-finger swipe. Fig.~\ref{fig-transparency:screenshot} shows a screenshot of the default view of the application. The top-right corner shows the menu buttons to toggle the visibility of the transparency modes. The bottom-left corner shows the real-time global coordinate frame.

\subsection{Granularity of Control}
In our previous work~\cite{patel2019, patel2019improving}, we proposed an interface capable of mixed granularity of control for single operators. The `granularity' refers to the possibility to interact at the robot-, team-, and environment-level. Robot- and team-level control allow the operator to send direct commands to individual robots or groups of them. With environment-level granularity, the operator indicates the \textit{desired effect} of a modification of the environment, e.g., moving an object, and the robots autonomously execute the action. As discussed in~\cite{patel2019, patel2019improving}, mixed granularity of control can outperform any individual level of control. We use these interface features in the present study.

\begin{figure}[t]
  \centering
  \begin{subfigure}[t]{0.23\textwidth}
    \includegraphics[width=\textwidth]{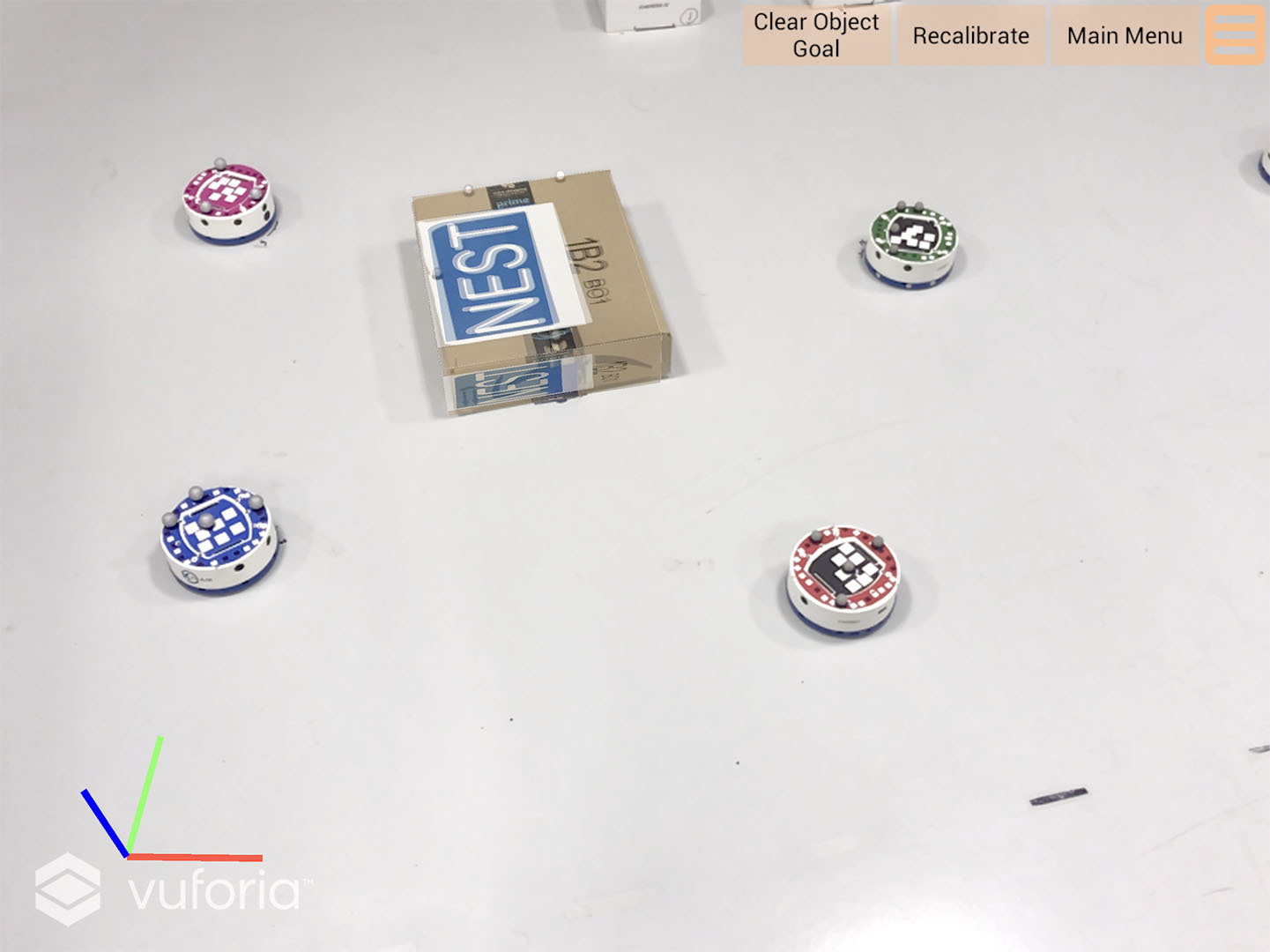}
    \caption{Object recognition}
    \label{fig-transparency:modeO1}
  \end{subfigure}
  \begin{subfigure}[t]{0.23\textwidth}
    \includegraphics[width=\textwidth]{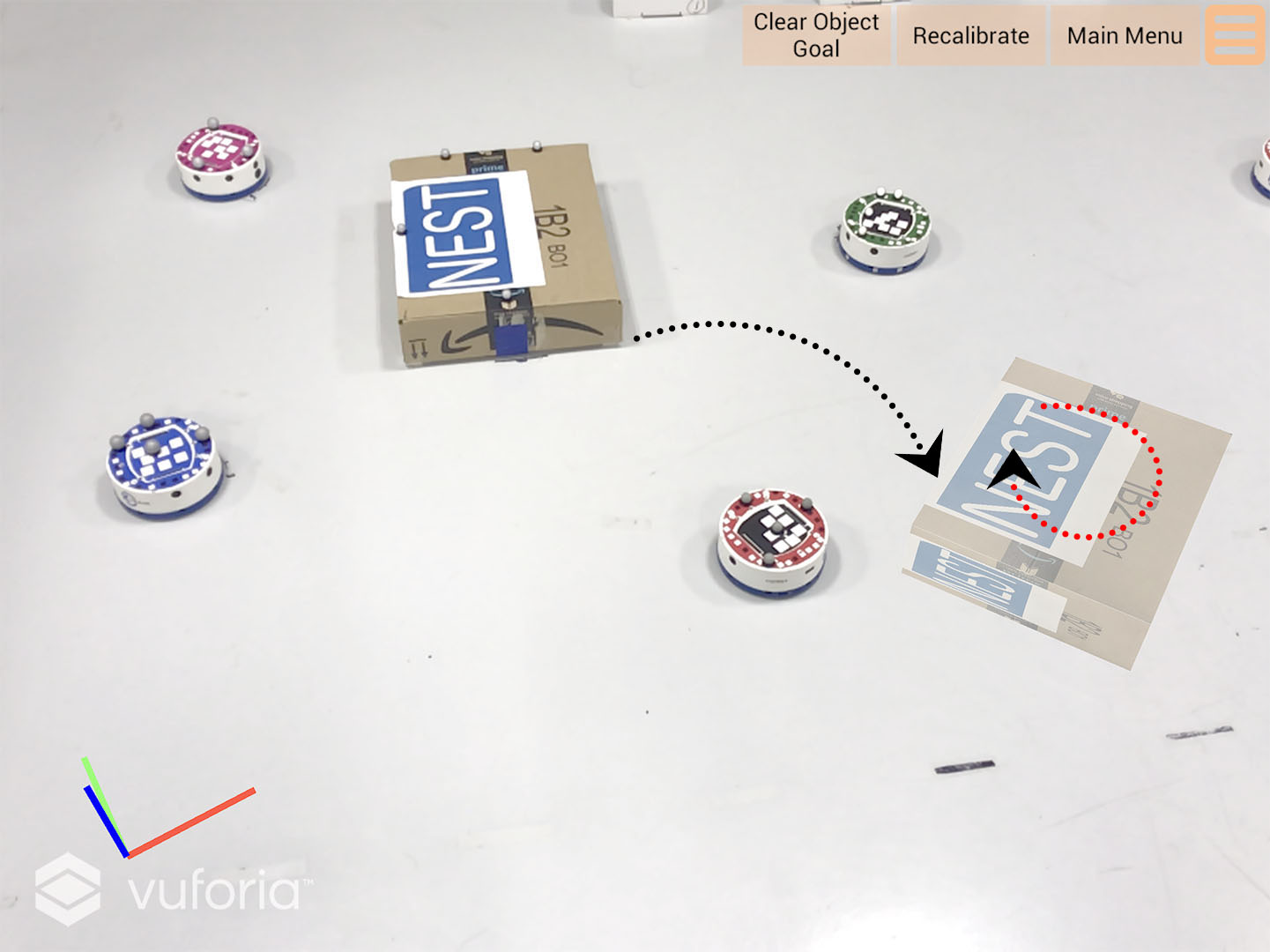}
    \caption{New Goal Defined}
    \label{fig-transparency:modeO2}
  \end{subfigure}
  \begin{subfigure}[t]{0.23\textwidth}
    \includegraphics[width=\textwidth]{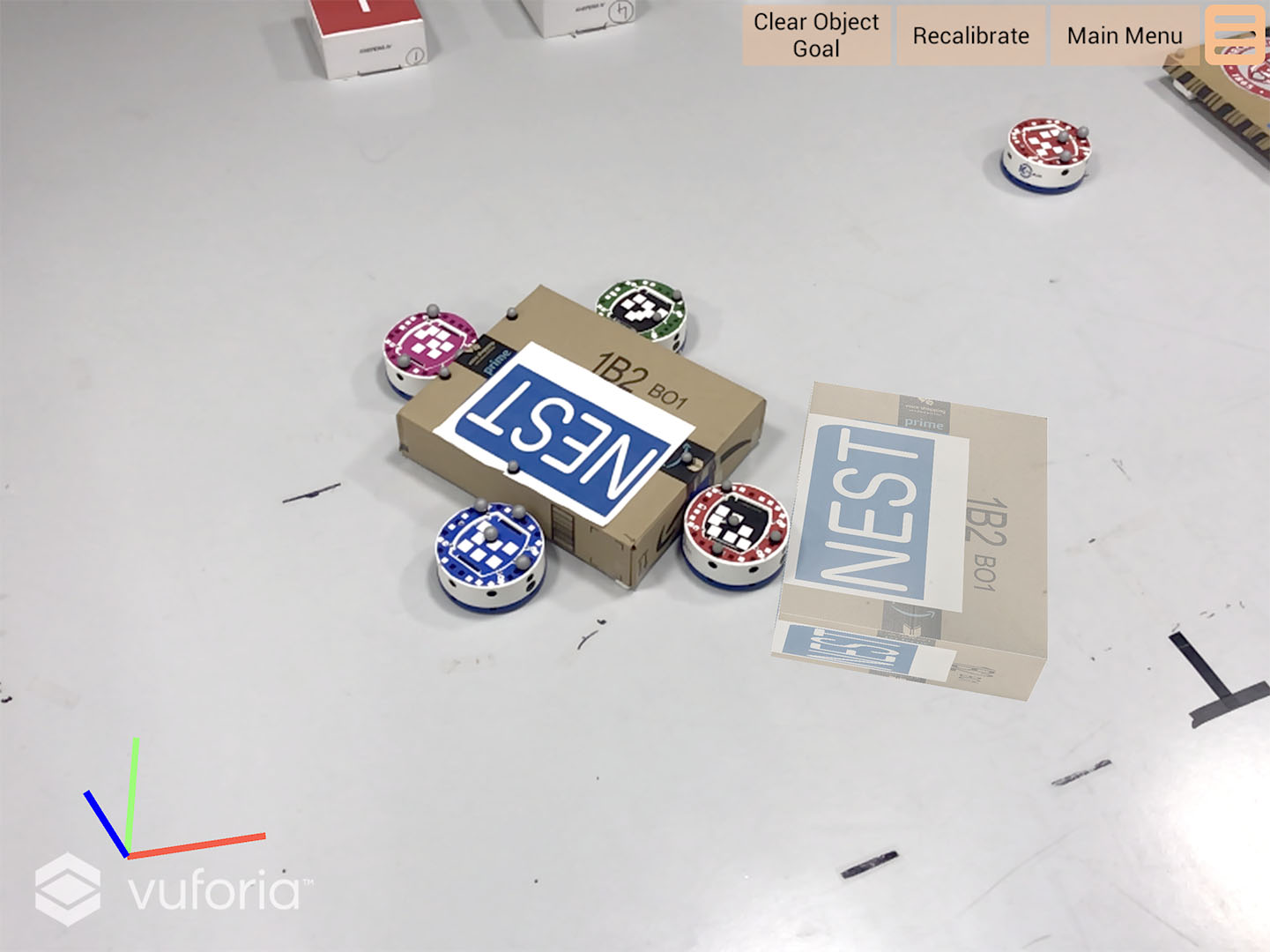}
    \caption{Robots approach and push}
    \label{fig-transparency:modeO3}
  \end{subfigure}
  \begin{subfigure}[t]{0.23\textwidth}
    \includegraphics[width=\textwidth]{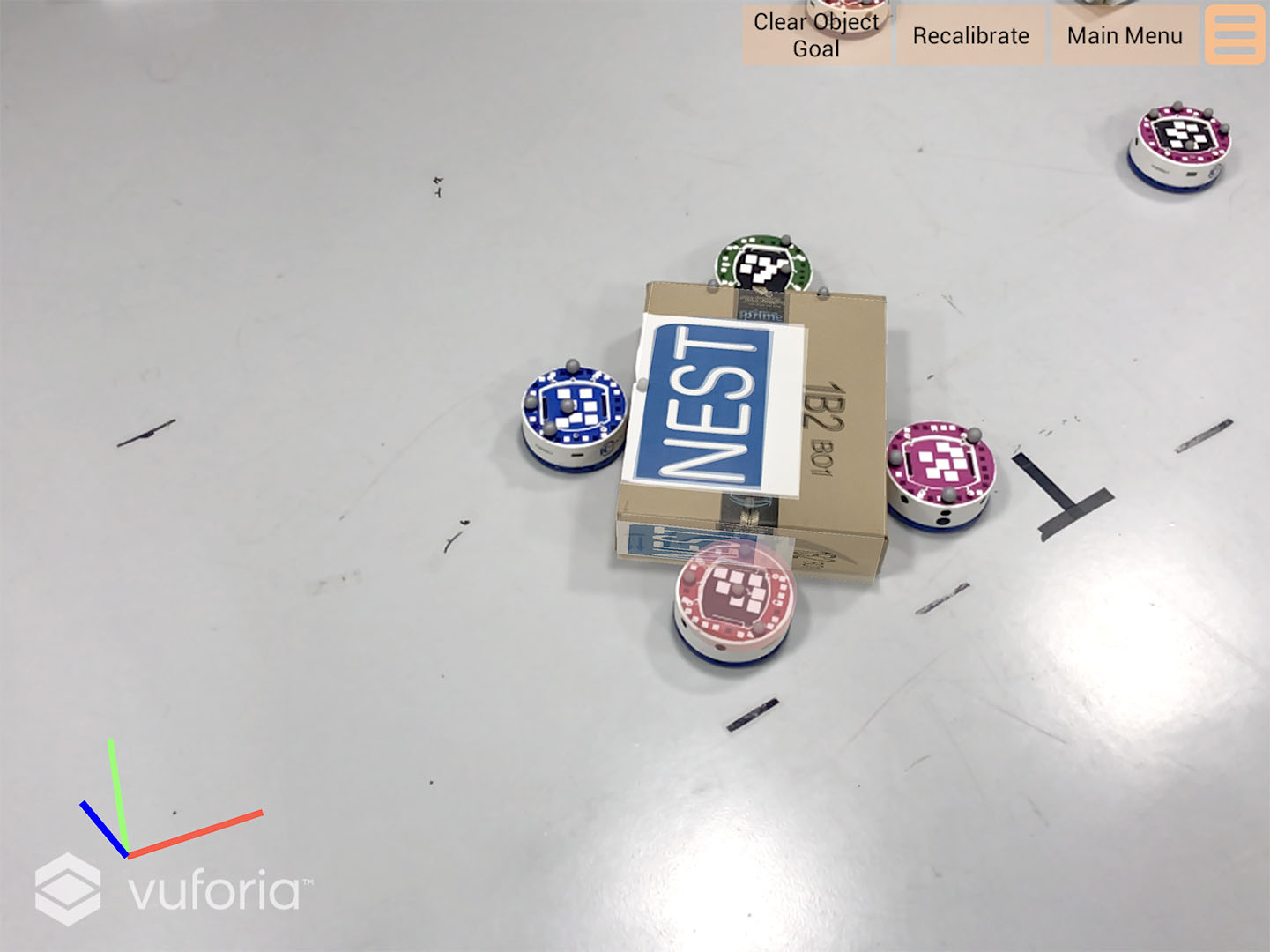}
    \caption{Transport complete}
    \label{fig-transparency:modeO4}
  \end{subfigure}
  \caption{Object manipulation by interaction with virtual objects. The overlaid dotted black arrow indicates the one-finger swipe gesture used to move the virtual object and the overlaid red dotted arrow indicates the two-finger rotation gesture.}\label{fig-transparency:modeO}
\end{figure}

\textbf{Object Manipulation.} The interface overlays virtual objects over the recognized objects (see Fig.~\ref{fig-transparency:modeO}). The user can move multiple virtual objects to define their respective desired poses, and teams of robots transport these objects to destination. If two or more operators simultaneously control the same object, the system processes the pose received last.

\begin{figure}[t]
  \centering
  \begin{subfigure}[t]{0.23\textwidth}
    \includegraphics[width=\textwidth]{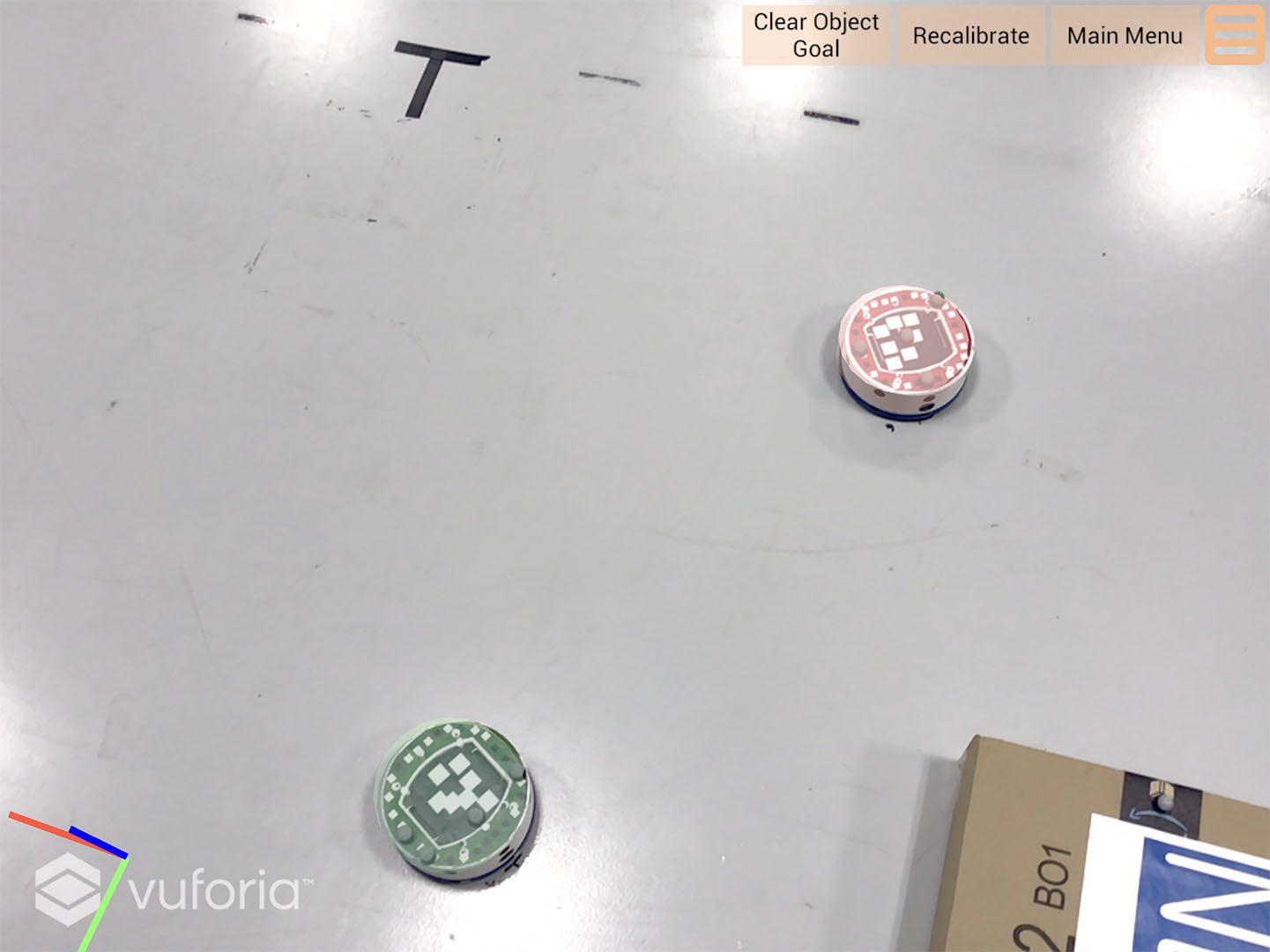}
    \caption{Robot recognition}
    \label{fig-transparency:modeR1}
  \end{subfigure}
  \begin{subfigure}[t]{0.23\textwidth}
    \includegraphics[width=\textwidth]{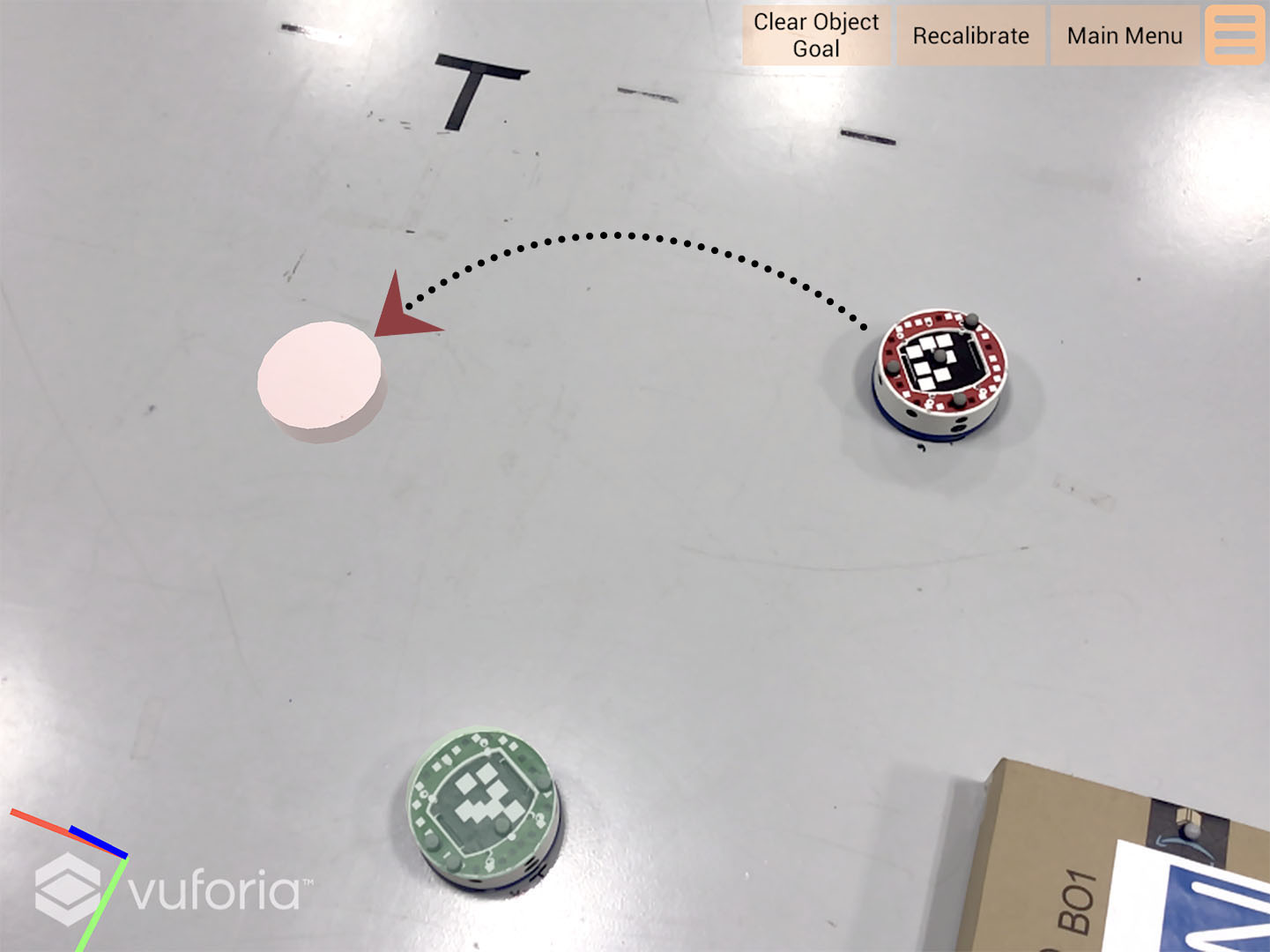}
    \caption{New robot position}
    \label{fig-transparency:modeR2}
  \end{subfigure}
  \caption{Robot manipulation by interaction with virtual robots. The overlaid dotted black arrow indicates the one-finger swipe gesture to move the virtual robot and the arrowhead color indicates the moved virtual robots.}\label{fig-transparency:modeR}
\end{figure} 

\textbf{Robot Manipulation.} The interface overlays virtual robots over the recognized robots (see Fig.~\ref{fig-transparency:modeR}). The color of the virtual robot resembles the color of the fiducial markers to differentiate between robots. The user can move multiple virtual robots to define their respective desired poses. If the robot is part of a team performing collective transport, the other robots in the same team pause until the selected robot reaches the desired pose. If the robot is part of a team not involved in collective transport, the selected robot overwrites the goal pose with the newly defined pose and does not affect its team members. If two or more operators simultaneously control the same robot, the system processes the last pose received. Fig.~\ref{fig-transparency:modeR} shows how virtual robots look.

\begin{figure}[t]
  \centering
  \begin{subfigure}[t]{0.23\textwidth}
    \includegraphics[width=\textwidth]{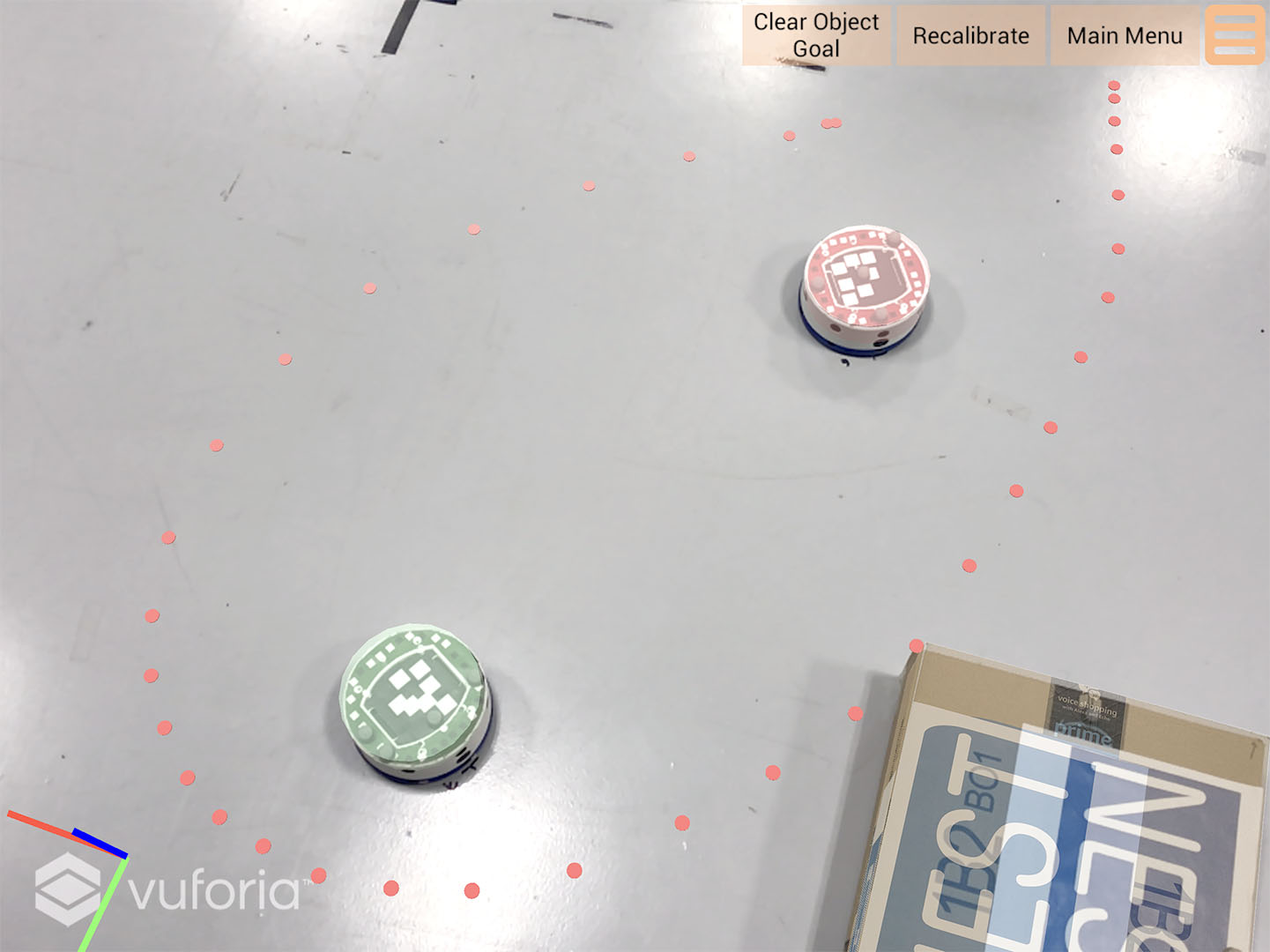}
    \caption{Robot team selection}
    \label{fig-transparency:modeS1}
  \end{subfigure}
  \begin{subfigure}[t]{0.23\textwidth}
    \includegraphics[width=\textwidth]{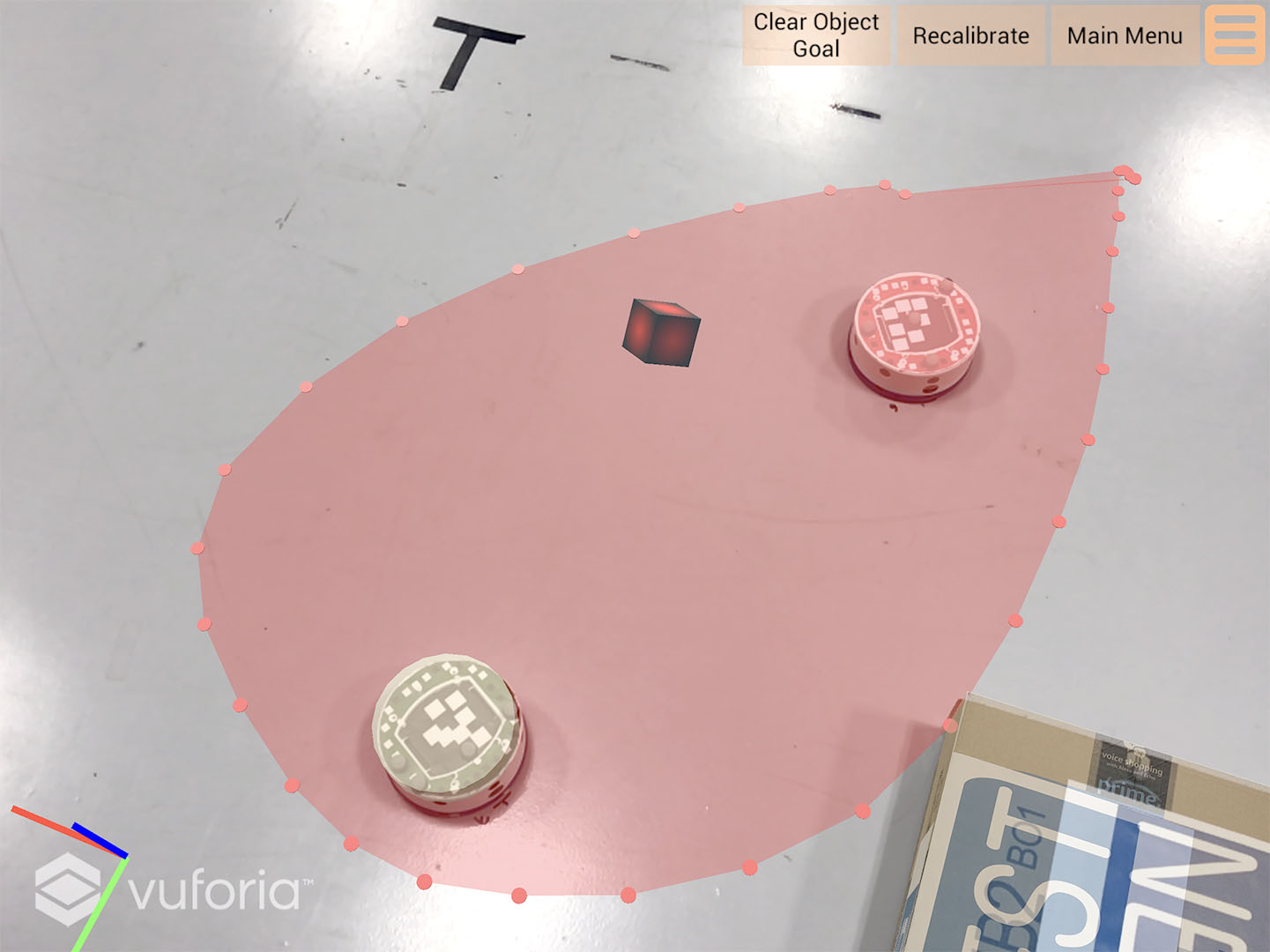}
    \caption{Robot team creation}
    \label{fig-transparency:modeS2}
  \end{subfigure}
  \begin{subfigure}[t]{0.23\textwidth}
    \includegraphics[width=\textwidth]{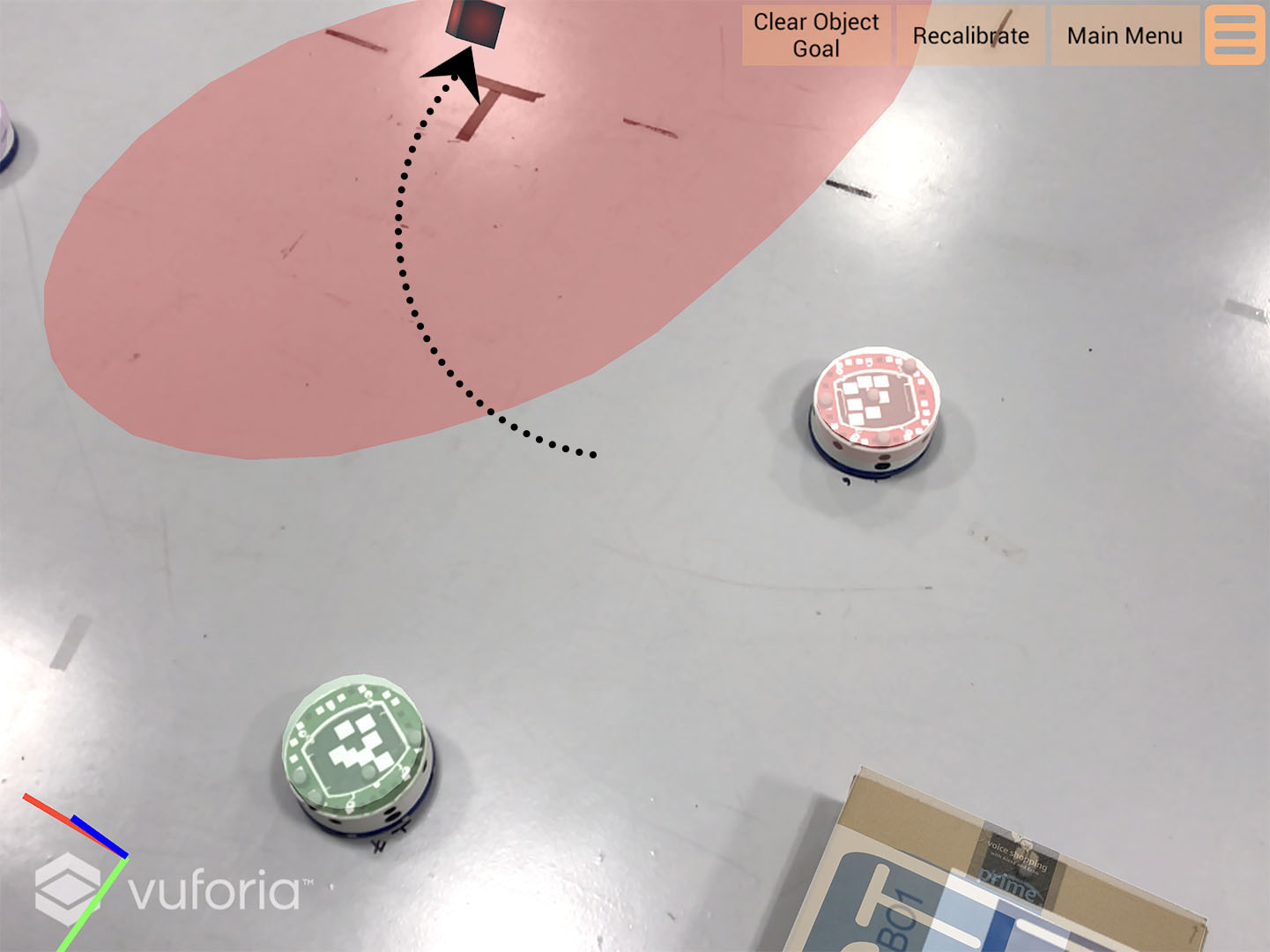}
    \caption{Robot team manipulation}
    \label{fig-transparency:modeS3}
  \end{subfigure}
  \begin{subfigure}[t]{0.23\textwidth}
    \includegraphics[width=\textwidth]{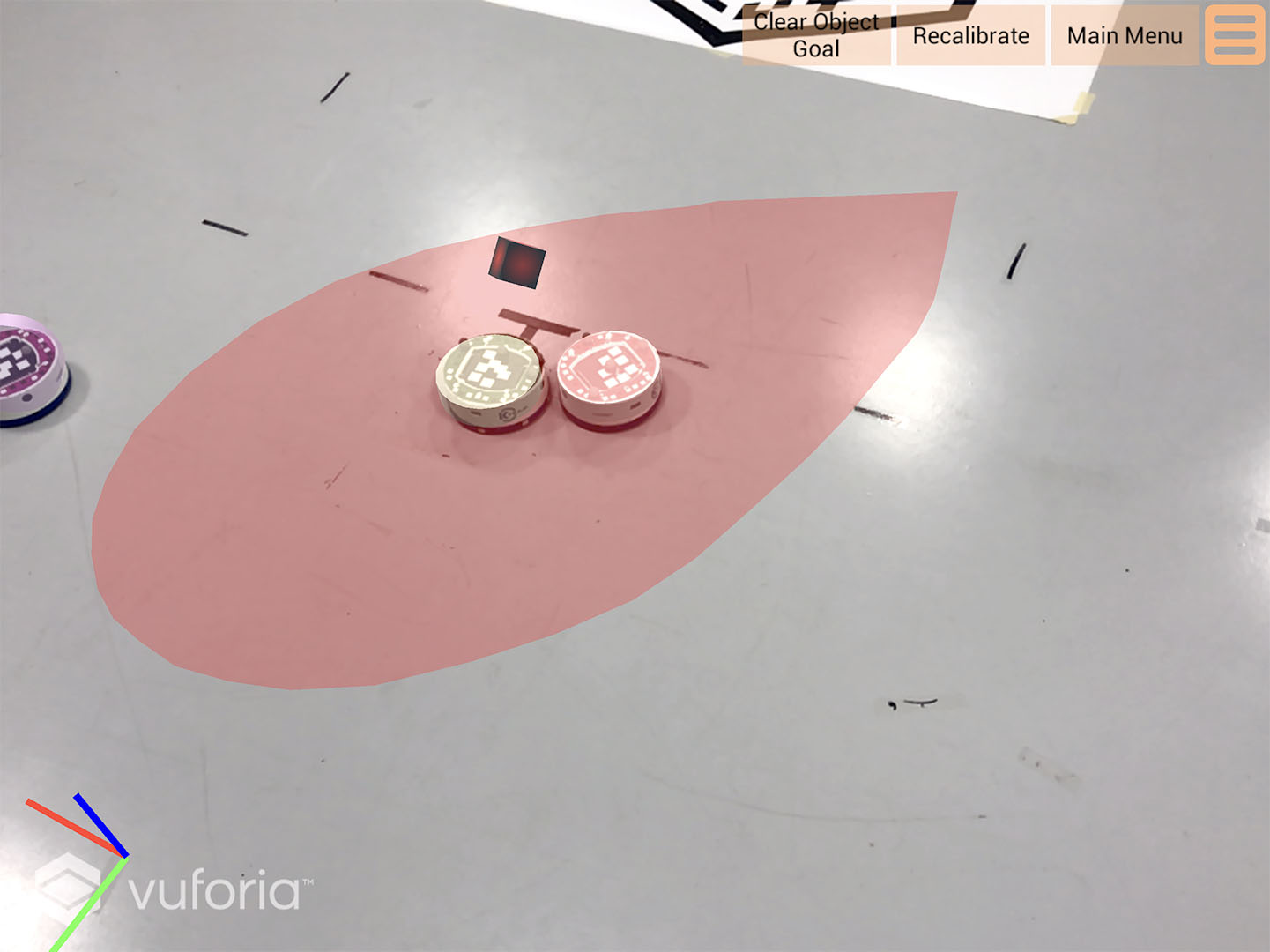}
    \caption{Robot team re-positioned}
    \label{fig-transparency:modeS4}
  \end{subfigure}
  \caption{Robot team creation and manipulation by interacting with the
    interface. The overlaid dotted black arrow indicates the one-finger swipe
    gesture to move the virtual cube for re-positioning the the team of
    robots.}\label{fig-transparency:modeS}
\end{figure} 

\textbf{Robot Team Selection and Manipulation.} The user can draw a closed path with a one-finger continuous swipe to select all the robots in the enclosed region (see Fig.~\ref{fig-transparency:modeS}). A contour-shaped virtual object with a virtual cube at its centroid appears in the graphical view. The user can manipulate this cube to define the desired pose for the selected team of robots. The user can handle only one team at a time. If two or more operators have the same robot in their team, the robot processes the pose received last. 

\subsection{Collective Transport}

\begin{figure}[t]
    \centering
    \includegraphics[width=0.7\linewidth]{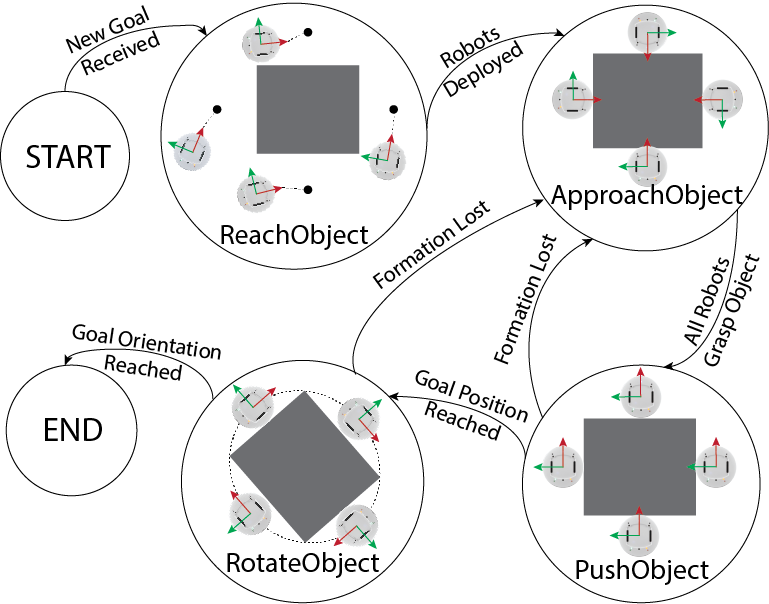}
    \caption{Collective transport state machine.}
    \label{fig-transparency:collective_transport}
\end{figure}

We employ a collective transport behavior based on the finite state
machine (FSM) shown in Fig.~\ref{fig-transparency:collective_transport}. This behavior is identical to the one presented in our previous work~\cite{patel2019}. The states in the FSM are explained next.

\textbf{Reach Object.} Upon receiving the desired goal pose for the object, the robots organize themselves around the object in a circular manner. These poses are decided based on the number of robots in the team and their distance from the object. This state comes to an end
when all the robots reach their designated poses.

\textbf{Approach Object.} The robots move towards the centroid of the object. This state is completed when all the robots are in contact with the object.

\textbf{Push Object.} The robots first rotate in-place facing the direction of the goal. The robots then move towards the goal. The robots modulate their speeds to maintain a set distance from the centroid of the object and keep their formation. If a robot breaks the formation, the team switches back to \textit{Approach Object}, waits for its completion, and subsequently resumes the transport behavior. The state comes to an end once the object reaches the goal position.

\textbf{Rotate Object.} The robots rearrange around the object and move along a circular path, thereby rotating the object in place. If any robot breaks the formation, the team rearranges and resumes object rotation. The state ends when the object reaches the desired orientation.

\subsection{Transparency Modes}

We present different transparency modes based on the visual FoV of our interface. The interface provides an option to switch between modes. The modes incorporate transparency features that reflect an operator's perception, comprehension, and projection, i.e., in terms of the three levels described in the SAT model~\cite{chen_situation_2014}. Table~\ref{tab-transparency:SATanalogy} lists the features, the transparency modes, and the corresponding the SAT levels of information.

\begin{table}[t]
\centering
\caption{Analogy between the features in our interface and the levels of the SAT model.}
\renewcommand{\arraystretch}{1.3}
\footnotesize
\begin{tabular}{p{2.7cm}|c|c}
\hline\hline
Transparency mode                       & Our feature               & SAT level         \\ \hline\hline
\multirow{3}{*}{Central Transparency}   & On-robot status           & Level 1 + 2       \\ 
                                        & Robot Direction Pointer   & Level 2           \\  
                                        & Shared Awareness          & Level 3           \\ \hline   
\multirow{3}{*}{Peripheral Transparency}& On-robot status  & Level 1 + 2       \\
                                        & Object Panel     & Level 2           \\
                                        & Text-based Log   & Level 3           \\ \hline                                       
\end{tabular}
\label{tab-transparency:SATanalogy}
\renewcommand{\arraystretch}{1}
\end{table}

\textbf{No Transparency (NT).} Operators can send control commands, but without access to any feedback information.

\begin{figure}[t]
    \centering
    \includegraphics[width=0.7\linewidth]{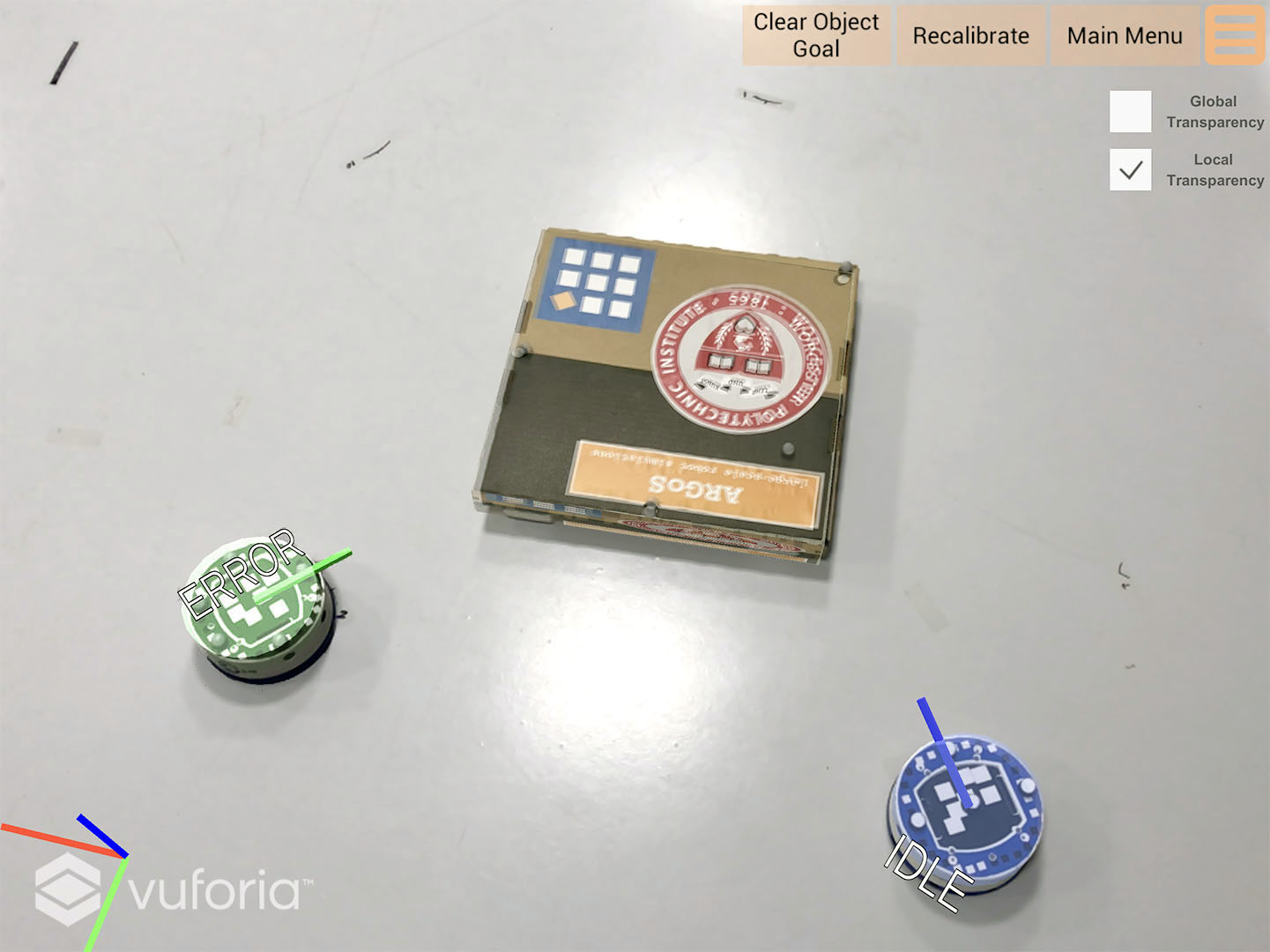}
    \caption{Central Transparency showing on-robot status and directional indicator.}
    \label{fig-transparency:central}
\end{figure}

\textbf{Central Transparency (CT).} The interface overlays each robot with a direction vector and text to report the current task (see Fig.~\ref{fig-transparency:central}). The direction vector indicates the heading of the robot. The color of the vectors resemble the color of the fiducial markers to differentiate between vectors when the robots are close to each other. The interface updates the information 10 times per second. The displayed states are: \textit{Idle}, \textit{Reach}, and \textit{Error}. The interface also reports the commands of other operators in real time, to foster collaboration and shared awareness, and to minimize (ideally avoid) conflicting control of the same robots and objects. This information is only visible if an operator is focusing the tablet camera on a specific robot or object, i.e., at the center of the FoV.

\begin{figure}[t]
    \centering
    \includegraphics[width=0.7\linewidth]{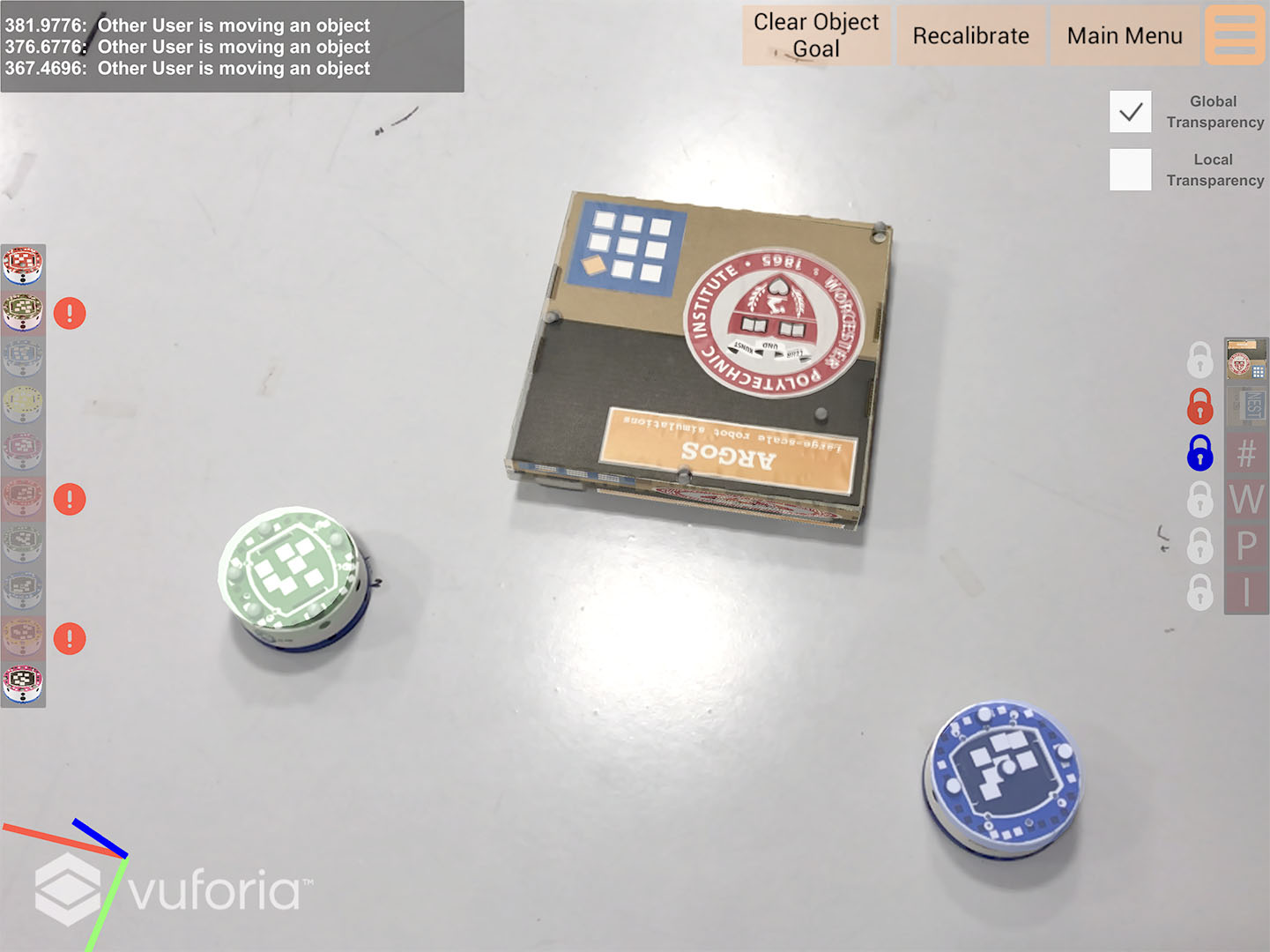}
    \caption{Peripheral Transparency mode showing text-based lock, object panel and robot panel (clockwise from top-left).}
    \label{fig-transparency:peripheral}
\end{figure}

\textbf{Peripheral Transparency (PT).} The interface displays a robot panel, an object panel, and a text-based log at the edges of the screen (see Fig.~\ref{fig-transparency:peripheral}). The robot panel shows the robots as icons. The highlighted icons correspond to the robots that are moving or performing operator-defined actions. The interface conveys error conditions as blinking red exclamation points. Analogously, the object panel shows the objects as icons. The interface highlights the icons that correspond to an object manipulated by the robots. The interface also offers the option to select an object icon to lock it for future use. By locking, an operator indicates that they intend to work with that object. The interface of other operators highlights the lock with a red icon. An operator can lock only one object at a time, removing past locks when a new one is requested. The text-based log reports the last 3 control actions taken by other operators. 

\textbf{Mixed Transparency (MT).} This mode offers the features of both central and peripheral transparency.

\section{User Study} \label{sec-transparency:userstudy}
\subsection{Hypotheses}
The primary purpose of this work is to investigate the effect of different transparency modes on the operators' awareness, workload, trust, interaction, and performance in a multi-human multi-robot scenario. We based our experiments on three hypotheses:
\begin{enumerate}
\item[\textbf{H1:}] Mixed transparency (MT) has the best outcome as compared to other modes, in terms of the mentioned metrics.
\item[\textbf{H2:}] Operators prefer mixed transparency (MT) to the other modes.
\item[\textbf{H3:}] Operators prefer central transparency (CT) to peripheral transparency (PT).
\end{enumerate}

\subsection{Gamified User Study}

\begin{figure}[t]
    \centering
    \includegraphics[width=0.7\linewidth]{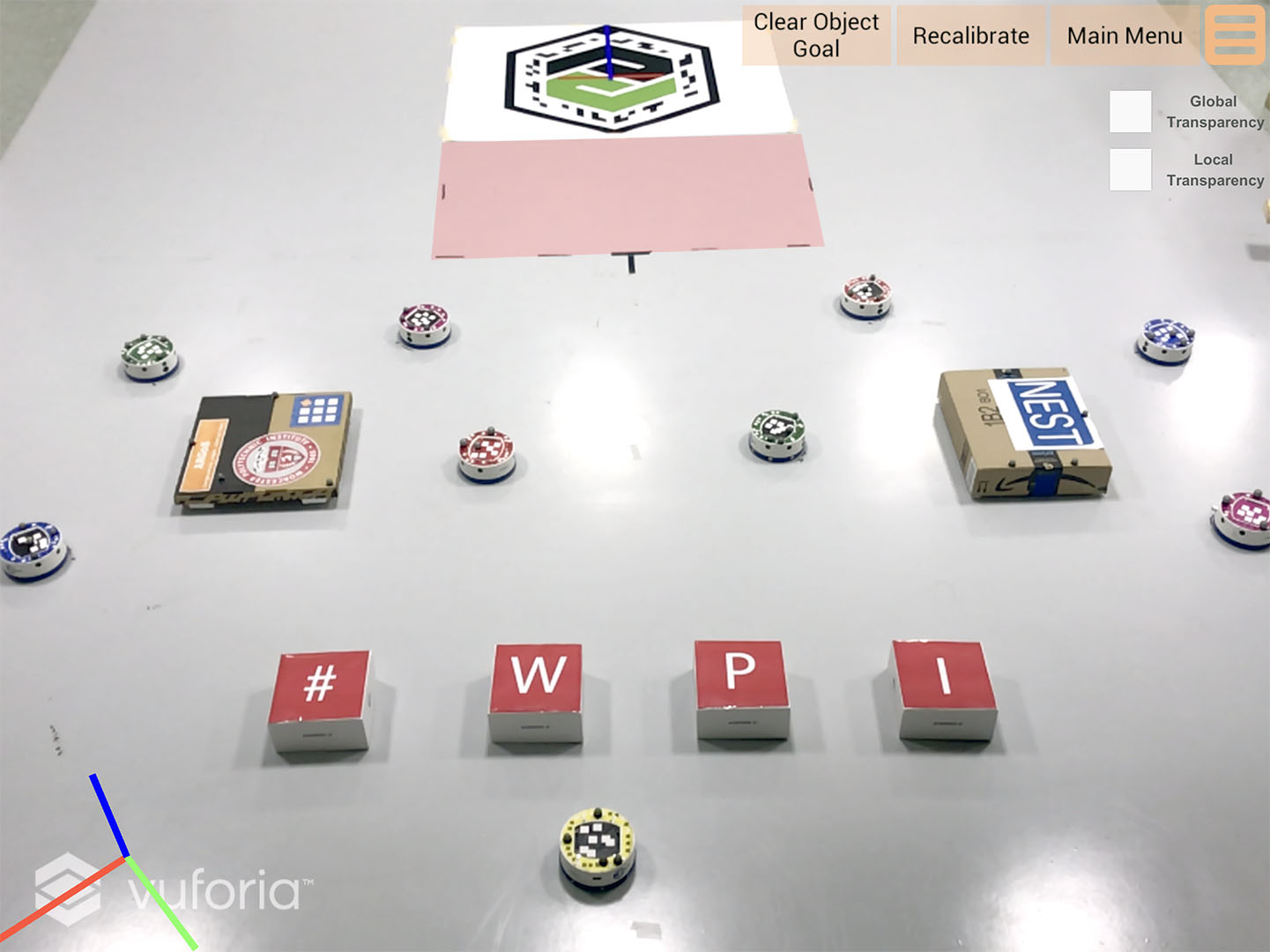}
    \caption{User study experiment setup. The shaded red box indicates the goal region.}
    \label{fig-transparency:study_setup}
\end{figure}

We devised a gamified scenario in which the operators must use the robots to perform object transport. Teams of two participants had to move 6 objects (2 big and 4 small) from their initial position to a goal region. Big objects were worth 2 points each, and small objects were worth 1 point each. The operators had to work collaboratively to gain as many points as possible (out of a maximum of 8) in a fixed time limit of 8 minutes. The operators could move the big objects using the collective transport behavior, or using the robot or robot-team manipulation modalities at will. Small objects could only be transported with the robot and team control modalities. The operators were given 9 robots to complete the game. Fig.~\ref{fig-transparency:study_setup} shows the initial positions of the robots, the objects, and the goal region.

\subsection{Participant Sample}
We recruited 18 university students (10 female, 8 male) with ages ranging from 19 to 41 ($23.78 \pm 5.08$) in accordance with protocols approved by WPI's IRB\footnote{https://www.wpi.edu/research/resources/compliance/institutional-review-board}. No participant had any prior experience with the system.

\subsection{Procedures}
Each session of the study approximately took 105 minutes and involved four games. After signing the consent form, we explained the scenario and gave the participants 10 minutes to play with the system. After each game, the participants had to answer a subjective questionnaire. All the participants played the game with all transparency modes (NT, CT, PT, MT) once. We randomized the order of the modes to reduce learning effects. 

\subsection{Metrics}
We recorded subjective and objective measures for each participant for each task. We used the following measures:

\textbf{Situational Awareness.} We used the Situational Awareness Rating Technique (SART)~\cite{taylor2017situational} on a 10-point Likert scale~\cite{likert} to assess the situational awareness after each game.

\textbf{Task Workload.} We used the NASA TLX~\cite{hart1988development} scale on a 4-point Likert scale to compare the perceived workload in each game.

\textbf{Trust.} We used the trust questionnaire~\cite{uggirala2004measurement} on a 10-point Likert scale to compare the trust in the interface affected by each transparency mode.

\textbf{Interaction.} We used a custom questionnaire (see Fig.~\ref{fig-transparency:questionnaire}) on a 5-point Likert scale to assess the operator-level and robot-level interaction.
\begin{figure}[t]
\centering
\footnotesize
\begin{framed}
\setdefaultleftmargin{0pt}{}{}{}{}{}
\begin{compactitem}
\item Did you understand your \textit{teammate’s intentions}? Were you able to understand why your teammate took a certain action?
\item Could you understand your \textit{teammate’s actions}? Could you understand what your teammate was doing at any particular time?
\item Could you follow the \textit{progress of the game}? Were you able to gauge how much of it was pending?
\item Did you understand what the \textit{robots were doing}, i.e., how and why the robots were behaving the way they did?
\item Was the information provided by the interface \textit{clear to understand}?
\end{compactitem}
\end{framed}
\caption{Questionnaire to assess the operator-level and robot-level interaction. Each question is expressed with a 5-point Likert scale.}
\label{fig-transparency:questionnaire}
\end{figure}

\textbf{Performance.} We used the points earned in each game as a metric to scale the performance achieved with each transparency mode.

\textbf{Usability.} We asked the participants to select the features (Log, Robot Panel, Object Panel, and On-Robot Status) they used during the study. Additionally, we asked them to rank the transparency modes from 1 to 4, 1 being the highest rank. 

\begin{table}[t]
\caption{Results with relationships between transparency modes. The relationship are based on mean ranks obtained through Friedman's Test. The symbol $^*$ denotes significant difference ($p<0.05$) and the symbol $^{**}$ denotes marginally significant difference ($p<0.10$). The symbol $^-$ denotes negative scales and lower ranking is a good ranking.}
\footnotesize
\renewcommand{\arraystretch}{1.3}
\begin{tabular}{c|c|c|c}
\hline
\textbf{Attributes}            & \textbf{Relationship}      & \textbf{$\chi^2(3)$}   & \textbf{$p$-value}    \\ \hline 
\multicolumn{4}{c}{\textbf{SART SUBJECTIVE SCALE}}                                                           \\ \hline\hline
Instability of Situation$^-$   & not significant            & $4.192$                 & $0.241$              \\ 
Complexity of Situation$^-$    & NT$>$MT$>$PT$>$CT$^{**}$   & $6.435$                 & $0.092$              \\
Variability of Situation$^-$   & not significant            & $4.192$                 & $0.241$              \\ 
Arousal                        & NT$>$MT$>$PT$>$CT$^{**}$   & $7.093$                 & $0.069$              \\
Concentration of Attention     & not significant            & $4.664$                 & $0.198$              \\
Spare Mental Capacity          & not significant            & $3.526$                 & $0.317$              \\
Information Quantity           & MT$>$CT$=$PT$>$NT$^{*}$    & $16.160$                & $0.001$              \\
Information Quality            & MT$>$CT$>$PT$>$NT$^{*}$    & $11.351$                & $0.010$              \\
Familiarity with Situation     & not significant            & $1.911$                 & $0.591$              \\ \hline\hline
\multicolumn{4}{c}{\textbf{NASA TLX SUBJECTIVE SCALE}}                                                       \\ \hline\hline
Mental Demand$^-$              & not significant            & $6.169$                 & $0.104$              \\ 
Physical Demand$^-$            & not significant            & $3.526$                 & $0.317$              \\ 
Temporal Demand$^-$            & not significant            & $0.564$                 & $0.903$              \\ 
Performance$^-$                & not significant            & $4.573$                 & $0.206$              \\ 
Effort$^-$                     & NT$>$PT$>$CT$>$MT$^{* }$   & $9.203$                 & $0.027$              \\ 
Frustration$^-$                & NT$>$CT$>$MT$>$PT$^{* }$   & $9.205$                 & $0.027$              \\ \hline\hline
\multicolumn{4}{c}{\textbf{TRUST SUBJECTIVE SCALE}}                                                          \\ \hline\hline
Competence                     & not significant            & $3.703$                 & $0.295$              \\ 
Predictability                 & PT$>$CT$>$MT$>$NT$^{**}$   & $6.359$                 & $0.095$              \\ 
Reliability                    & not significant            & $4.338$                 & $0.227$              \\ 
Faith                          & not significant            & $1.891$                 & $0.595$              \\ 
Overall Trust                  & PT$>$MT$>$CT$=$NT$^{* }$   & $12.607$                & $0.005$              \\ 
Accuracy                       & PT$>$MT$=$CT$>$NT$^{* }$   & $12.214$                & $0.007$              \\ \hline\hline
\multicolumn{4}{c}{\textbf{INTERACTION SUBJECTIVE SCALE}}                                                    \\ \hline\hline
Teammate's Intent              & MT$>$PT$>$CT$>$NT$^{*}$    & $23.976$                & $0.000$              \\ 
Teammate's Action              & MT$>$PT$=$CT$>$NT$^{*}$    & $22.511$                & $0.000$              \\ 
Task Progress                  & MT$>$CT$>$PT$>$NT$^{*}$    & $25.619$                & $0.000$              \\ 
Robot Status                   & CT$>$PT$>$MT$>$NT$^{*}$    & $13.608$                & $0.003$              \\ 
Information Clarity            & CT$>$PT$>$MT$>$NT$^{*}$    & $12.078$                & $0.007$              \\ \hline\hline
\multicolumn{4}{c}{\textbf{PERFORMANCE OBJECTIVE SCALE}}                                                     \\ \hline\hline
Points Scored                  & not significant            & $5.554$                 & $0.135$              \\ \hline 
\end{tabular}
\label{tab-transparency:results}
\renewcommand{\arraystretch}{1}
\end{table}

\begin{figure}[t]
    \centering
    \includegraphics[width=0.7\linewidth]{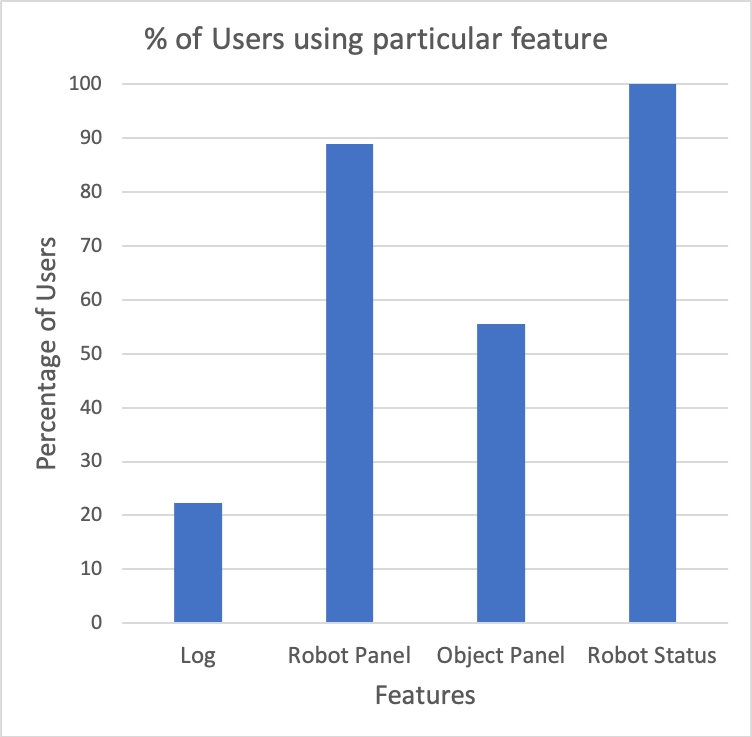}
    \caption{Feature Usability.}
    \label{fig-transparency:feature_usability}
\end{figure}

\begin{figure}[t]
    \centering
    \includegraphics[width=0.7\linewidth]{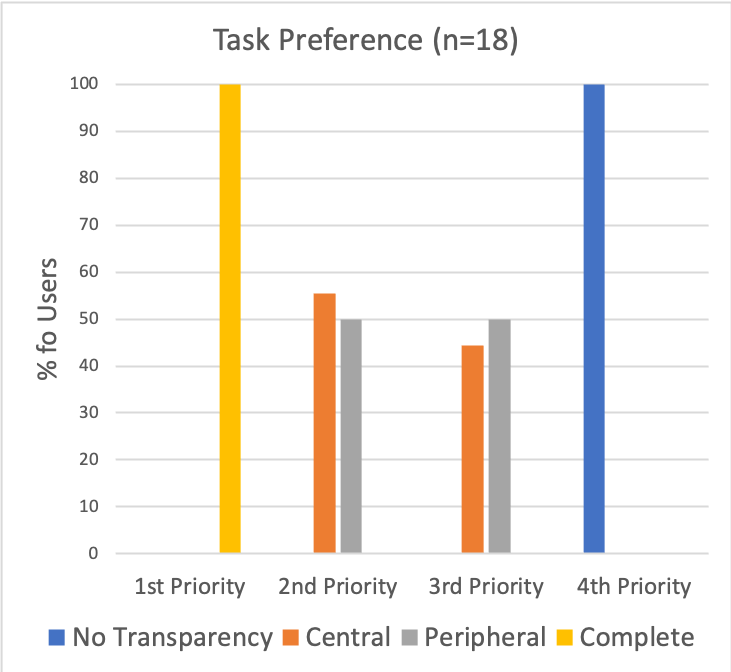}
    \caption{Game Preference.}
    \label{fig-transparency:task_usability}
\end{figure}

\begin{table}[t]
\centering
\caption{Ranking scores based on the Borda count. The gray cells indicate the leading scenario for each type of ranking.}
\renewcommand{\arraystretch}{1.3}
\footnotesize
\begin{tabular}{p{4cm}|c|c|c|c}
\hline\hline
Borda Count                                                         & NT   & CT    & PT    & MT                            \\ \hline\hline
Based on Collected Data Ranking (Table~\ref{tab-transparency:results})           & 17.5 & 40    & 39    & \cellcolor[HTML]{EFEFEF}43.5  \\  
Based on Preference Data Ranking (Fig.~\ref{fig-transparency:task_usability})    & 18   & 46    & 45    & \cellcolor[HTML]{EFEFEF}72    \\ \hline 
\end{tabular}
\label{tab-transparency:borda}
\renewcommand{\arraystretch}{1}
\end{table}

\section{Analysis and Discussion}
\label{sec-transparency:discussion}

Table~\ref{tab-transparency:results} shows the results for all the subjective scales and the objective metrics. We used the Friedman test~\cite{friedman1937use} to analyze the data and to assess the significance among different games. We formed rankings based on the mean ranks for all the attributes that showed statistical significance (set to $p<0.05$) or marginal significance (set to $p<0.10$). Fig.~\ref{fig-transparency:feature_usability} shows the percentage of operators using a particular feature. Fig.~\ref{fig-transparency:task_usability} reports how participants ranked the transparency modes.

We used the Borda count~\cite{black1976partial} to calculate the rankings. We inverted the ranking of the negative scales when calculating the Borda count scores. Table~\ref{tab-transparency:borda} shows the results for each category. This table indicates mixed transparency (MT) as the overall winner in terms of performance, as well as the preferred mode across participants, in accordance with hypotheses H1 and H2. The data suggests that central transparency is better than peripheral transparency, confirming hypothesis H3. 

\textbf{Mixed Transparency.} This mode is the overall best choice for operators. The data suggests that this mode has the best information quality and quantity. The operators could pick the information they wanted from central and peripheral regions. However, the operators reported higher perceived complexity and higher arousal than with other transparency modes. The usability tests suggested this mode as the best to understand the teammate's intent and actions. This justifies mixed transparency as the first choice.

\textbf{Central Transparency.} This mode has the lowest complexity and arousal. The users found it easier to focus on the center of the screen, and use on-robot status over the side panels. The users reported better information quality and clarity w.r.t to peripheral transparency. $55.55\%$ of the operators preferred this mode over peripheral transparency. 

\textbf{Peripheral Transparency.} The operators found the information displayed at periphery of the screen hard to parse and access. This led to increased effort, complexity, and arousal w.r.t. the central transparency mode. However, as the information was available on-demand and was not constantly displayed in the FoV, the users reported the lowest amount of frustration. Operators also preferred the icon panels over the text-based log. Additionally, the operators preferred PT over CT to gain awareness of their teammate's intention. 

\textbf{Performance.} Our experiments did not report a substantial difference in performance across transparency modes. We hypothesize that this lack of difference is due to a learning effect across the four runs that each team had to perform. We could not avoid this learning effect through randomization of the transparency modes or pre-study training. The training sessions improved the participants' understanding of the interface and its features, but did not improve operation proficiency. We attribute this issue to the fact that our study was conducted with real robots, exposing the participants to real-world issues with robots they never encountered before (e.g. noise, failures).

\begin{figure}[t]
    \centering
    \includegraphics[width=0.7\columnwidth]{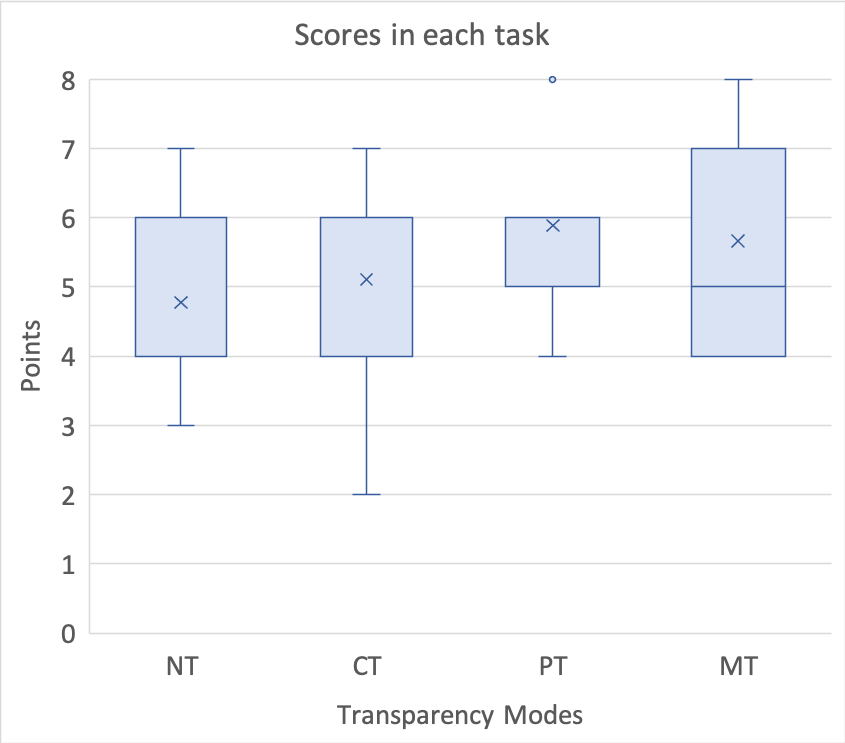}
    \caption{Task performance for each transparency mode.}
    \label{fig-transparency:performance}
\end{figure}

\begin{figure}[t]
    \centering
    \includegraphics[width=0.7\columnwidth]{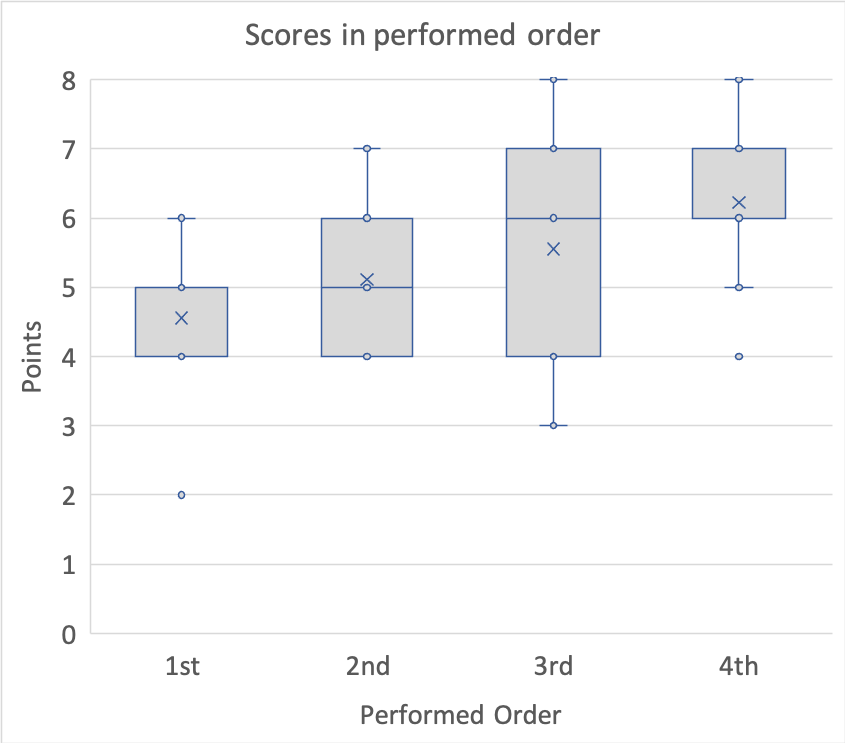}
    \caption{Learning effect in the user study.}
    \label{fig-transparency:learning}
\end{figure}

Fig.~\ref{fig-transparency:performance} reports the performance recorded in each game. Fig.~\ref{fig-transparency:learning} shows the increase in performance sorted by game performed (learning effect). Task performance dropped or stayed the same for teams that used no transparency after using other transparency modes. 

\section{Conclusion and Future Work} \label{sec-transparency:conclusion}
In this paper, we studied the effects of different transparency modes in multi-human multi-robot interaction. We classified transparency based on visual FoV. We demonstrated the design of a novel augmented-reality interface that supports different modes of transparency and provides both operator-level and robot-level information. 

We performed a user study with 18 operators to assess the effects of these modes of transparency on awareness, workload, trust, and interaction. Mixed transparency outperformed other modes in terms of overall effect and usability, and the participants chose mixed transparency as the best mode. We also compared central transparency with peripheral transparency. More operators preferred central transparency (55.55\%) over peripheral transparency (45.45\%). Although the difference between the central and peripheral transparency is small, these modes of transparency have their respective benefits. Central transparency offers better robot-level information, while peripheral transparency provides better operator-level information.

We recognize that the sample size of our study is limited, making our study in some ways exploratory from a statistical standpoint. However, the complexity of the task we studied is compelling, especially when compared with existing literature. The next iteration of our work will focus on expanding the user study in two directions. First, understanding the effects of learning and training on transparency, i.e., comparing the need of information for a novice user with the needs of an expert user. Second, studying the effects of our transparency features on the operator's reaction time, i.e., the time taken to resolve a problem.

\section*{Acknowledgements}
This work was funded by an Amazon Research Award.

\bibliographystyle{IEEEtran}
\bibliography{ref}

\begin{thebibliography}{10}
\providecommand{\url}[1]{#1}
\csname url@samestyle\endcsname
\providecommand{\newblock}{\relax}
\providecommand{\bibinfo}[2]{#2}
\providecommand{\BIBentrySTDinterwordspacing}{\spaceskip=0pt\relax}
\providecommand{\BIBentryALTinterwordstretchfactor}{4}
\providecommand{\BIBentryALTinterwordspacing}{\spaceskip=\fontdimen2\font plus
\BIBentryALTinterwordstretchfactor\fontdimen3\font minus
  \fontdimen4\font\relax}
\providecommand{\BIBforeignlanguage}[2]{{%
\expandafter\ifx\csname l@#1\endcsname\relax
\typeout{** WARNING: IEEEtran.bst: No hyphenation pattern has been}%
\typeout{** loaded for the language `#1'. Using the pattern for}%
\typeout{** the default language instead.}%
\else
\language=\csname l@#1\endcsname
\fi
#2}}
\providecommand{\BIBdecl}{\relax}
\BIBdecl

\bibitem{Brambilla2013}
M.~Brambilla, E.~Ferrante, M.~Birattari, and M.~Dorigo, ``{Swarm robotics: A
  review from the swarm engineering perspective},'' \emph{Swarm Intelligence},
  vol.~7, no.~1, pp. 1--41, 2013.

\bibitem{murphy2014disaster}
R.~R. Murphy, \emph{Disaster robotics}.\hskip 1em plus 0.5em minus 0.4em\relax
  MIT press, 2014.

\bibitem{hamins2015research}
A.~P. Hamins, C.~Grant, N.~P. Bryner, A.~W. Jones, and G.~H. Koepke, ``Research
  roadmap for smart fire fighting,'' {NIST}, Tech. Rep. 1191, 2015.

\bibitem{goldsmith1999book}
D.~Goldsmith, ``Book review: Voyage to the milky way: the future of space
  exploration/tv books, 1999,'' \emph{Sky and Telescope}, vol.~98, no.~5,
  p.~81, 1999.

\bibitem{denkenberger2007comparison}
J.~S. Denkenberger, C.~T. Driscoll, S.~W. Effler, D.~M. O'Donnell, and D.~A.
  Matthews, ``Comparison of an urban lake targeted for rehabilitation and a
  reference lake based on robotic monitoring,'' \emph{Lake and Reservoir
  Management}, vol.~23, no.~1, pp. 11--26, 2007.

\bibitem{buters2019methodological}
T.~M. Buters, P.~W. Bateman, T.~Robinson, D.~Belton, K.~W. Dixon, and A.~T.
  Cross, ``Methodological ambiguity and inconsistency constrain unmanned aerial
  vehicles as a silver bullet for monitoring ecological restoration,''
  \emph{Remote Sensing}, vol.~11, no.~10, p. 1180, 2019.

\bibitem{rubio2012mining}
R.~F. Rubio, ``Mining: the challenge knocks on our door,'' \emph{Mine Water and
  the Environment}, vol.~31, no.~1, pp. 69--73, 2012.

\bibitem{oh2009bridge}
J.-K. Oh, G.~Jang, S.~Oh, J.~H. Lee, B.-J. Yi, Y.~S. Moon, J.~S. Lee, and
  Y.~Choi, ``Bridge inspection robot system with machine vision,''
  \emph{Automation in Construction}, vol.~18, no.~7, pp. 929--941, 2009.

\bibitem{sirouspour2005multi}
S.~Sirouspour and P.~Setoodeh, ``Multi-operator/multi-robot teleoperation: an
  adaptive nonlinear control approach,'' in \emph{2005 IEEE/RSJ International
  Conference on Intelligent Robots and Systems}.\hskip 1em plus 0.5em minus
  0.4em\relax IEEE, 2005, pp. 1576--1581.

\bibitem{roundtree_transparency:_2019}
\BIBentryALTinterwordspacing
K.~A. Roundtree, M.~A. Goodrich, and J.~A. Adams,
  ``\BIBforeignlanguage{en}{Transparency: {Transitioning} {From}
  {Human}–{Machine} {Systems} to {Human}-{Swarm} {Systems}},''
  \emph{\BIBforeignlanguage{en}{Journal of Cognitive Engineering and Decision
  Making}}, p. 155534341984277, Apr. 2019. [Online]. Available:
  \url{http://journals.sagepub.com/doi/10.1177/1555343419842776}
\BIBentrySTDinterwordspacing

\bibitem{bhaskara_agent_2020}
\BIBentryALTinterwordspacing
A.~Bhaskara, M.~Skinner, and S.~Loft, ``\BIBforeignlanguage{en}{Agent
  {Transparency}: {A} {Review} of {Current} {Theory} and {Evidence}},''
  \emph{\BIBforeignlanguage{en}{IEEE Transactions on Human-Machine Systems}},
  pp. 1--10, 2020. [Online]. Available:
  \url{https://ieeexplore.ieee.org/document/8982042/}
\BIBentrySTDinterwordspacing

\bibitem{chakraborti_explicability?_nodate}
T.~Chakraborti, A.~Kulkarni, S.~Sreedharan, D.~E. Smith, and S.~Kambhampati,
  ``\BIBforeignlanguage{en}{Explicability? {Legibility}? {Predictability}?
  {Transparency}? {Privacy}? {Security}? {The} {Emerging} {Landscape} of
  {Interpretable} {Agent} {Behavior}},''
  \emph{\BIBforeignlanguage{en}{Proceedings of the international conference on
  automated planning and scheduling}}, vol.~29, pp. 86--96, 2019.

\bibitem{chen_situation_2018}
\BIBentryALTinterwordspacing
J.~Y.~C. Chen, S.~G. Lakhmani, K.~Stowers, A.~R. Selkowitz, J.~L. Wright, and
  M.~Barnes, ``\BIBforeignlanguage{en}{Situation awareness-based agent
  transparency and human-autonomy teaming effectiveness},''
  \emph{\BIBforeignlanguage{en}{Theoretical Issues in Ergonomics Science}},
  vol.~19, no.~3, pp. 259--282, May 2018. [Online]. Available:
  \url{https://www.tandfonline.com/doi/full/10.1080/1463922X.2017.1315750}
\BIBentrySTDinterwordspacing

\bibitem{tulli_eects_nodate}
S.~Tulli, F.~Correia, S.~Mascarenhas, F.~S. Melo, and A.~Paiva,
  ``\BIBforeignlanguage{en}{Effects of {Agents}’ {Transparency} on
  {Teamwork}?}'' \emph{\BIBforeignlanguage{en}{International Workshop on
  Explainable, Transparent Autonomous Agents and Multi-Agent Systems}}, pp.
  22--37, 2019.

\bibitem{kalpagam_ganesan_better_2018}
\BIBentryALTinterwordspacing
R.~Kalpagam~Ganesan, Y.~K. Rathore, H.~M. Ross, and H.~Ben~Amor,
  ``\BIBforeignlanguage{en}{Better {Teaming} {Through} {Visual} {Cues}: {How}
  {Projecting} {Imagery} in a {Workspace} {Can} {Improve} {Human}-{Robot}
  {Collaboration}},'' \emph{\BIBforeignlanguage{en}{IEEE Robotics \& Automation
  Magazine}}, vol.~25, no.~2, pp. 59--71, Jun. 2018. [Online]. Available:
  \url{https://ieeexplore.ieee.org/document/8359206/}
\BIBentrySTDinterwordspacing

\bibitem{de_paolis_debugging_2019}
\BIBentryALTinterwordspacing
B.~Hoppenstedt, T.~Witte, J.~Ruof, K.~Kammerer, M.~Tichy, M.~Reichert, and
  R.~Pryss, ``\BIBforeignlanguage{en}{Debugging {Quadrocopter} {Trajectories}
  in {Mixed} {Reality}},'' in \emph{\BIBforeignlanguage{en}{Augmented
  {Reality}, {Virtual} {Reality}, and {Computer} {Graphics}}}, L.~T. De~Paolis
  and P.~Bourdot, Eds.\hskip 1em plus 0.5em minus 0.4em\relax Cham: Springer
  International Publishing, 2019, vol. 11614, pp. 43--50. [Online]. Available:
  \url{http://link.springer.com/10.1007/978-3-030-25999-0_4}
\BIBentrySTDinterwordspacing

\bibitem{lyons2013being}
J.~B. Lyons, ``Being transparent about transparency: A model for human-robot
  interaction,'' in \emph{2013 AAAI Spring Symposium Series}, 2013.

\bibitem{chen_situation_2014}
\BIBentryALTinterwordspacing
J.~Y. Chen, K.~Procci, M.~Boyce, J.~Wright, A.~Garcia, and M.~Barnes,
  ``\BIBforeignlanguage{en}{Situation {Awareness}-{Based} {Agent}
  {Transparency}:},'' Defense Technical Information Center, Fort Belvoir, VA,
  Tech. Rep., Apr. 2014. [Online]. Available:
  \url{http://www.dtic.mil/docs/citations/ADA600351}
\BIBentrySTDinterwordspacing

\bibitem{miller1956magical}
G.~A. Miller, ``The magical number seven, plus or minus two: Some limits on our
  capacity for processing information.'' \emph{Psychological review}, vol.~63,
  no.~2, p.~81, 1956.

\bibitem{lewis2010choosing}
M.~Lewis, H.~Wang, S.~Y. Chien, P.~Velagapudi, P.~Scerri, and K.~Sycara,
  ``Choosing autonomy modes for multirobot search,'' \emph{Human Factors},
  vol.~52, no.~2, pp. 225--233, 2010.

\bibitem{patel2019}
J.~Patel, Y.~Xu, and C.~Pinciroli, ``Mixed-granularity human-swarm
  interaction,'' in \emph{Robotics and {Automation} ({ICRA}), 2019 {IEEE}
  {International} {Conference} on}.\hskip 1em plus 0.5em minus 0.4em\relax
  IEEE, 2019.

\bibitem{patel2019improving}
J.~Patel and C.~Pinciroli, ``Improving human performance using mixed
  granularity of control in multi-human multi-robot interaction,'' \emph{arXiv
  preprint arXiv:1909.07487}, 2019.

\bibitem{schmorrow_proposed_2016}
\BIBentryALTinterwordspacing
S.~Lakhmani, J.~Abich, D.~Barber, and J.~Chen, ``\BIBforeignlanguage{en}{A
  {Proposed} {Approach} for {Determining} the {Influence} of {Multimodal}
  {Robot}-of-{Human} {Transparency} {Information} on {Human}-{Agent}
  {Teams}},'' in \emph{\BIBforeignlanguage{en}{Foundations of {Augmented}
  {Cognition}: {Neuroergonomics} and {Operational} {Neuroscience}}}, D.~D.
  Schmorrow and C.~M. Fidopiastis, Eds.\hskip 1em plus 0.5em minus 0.4em\relax
  Cham: Springer International Publishing, 2016, vol. 9744, pp. 296--307.
  [Online]. Available:
  \url{http://link.springer.com/10.1007/978-3-319-39952-2_29}
\BIBentrySTDinterwordspacing

\bibitem{felzmann_robots_2019}
\BIBentryALTinterwordspacing
H.~Felzmann, E.~Fosch-Villaronga, C.~Lutz, and A.~Tamo-Larrieux,
  ``\BIBforeignlanguage{en}{Robots and {Transparency}: {The} {Multiple}
  {Dimensions} of {Transparency} in the {Context} of {Robot} {Technologies}},''
  \emph{\BIBforeignlanguage{en}{IEEE Robotics \& Automation Magazine}},
  vol.~26, no.~2, pp. 71--78, Jun. 2019. [Online]. Available:
  \url{https://ieeexplore.ieee.org/document/8684252/}
\BIBentrySTDinterwordspacing

\bibitem{chien_influence_2019}
\BIBentryALTinterwordspacing
S.-Y. Chien, M.~Lewis, K.~Sycara, A.~Kumru, and J.-S. Liu,
  ``\BIBforeignlanguage{en}{Influence of {Culture}, {Transparency}, {Trust},
  and {Degree} of {Automation} on {Automation} {Use}},''
  \emph{\BIBforeignlanguage{en}{IEEE Transactions on Human-Machine Systems}},
  pp. 1--10, 2019. [Online]. Available:
  \url{https://ieeexplore.ieee.org/document/8836099/}
\BIBentrySTDinterwordspacing

\bibitem{zhu_enhancing_2020}
\BIBentryALTinterwordspacing
Y.~Zhu, T.~Aoyama, and Y.~Hasegawa, ``\BIBforeignlanguage{en}{Enhancing the
  {Transparency} by {Onomatopoeia} for {Passivity}-{Based} {Time}-{Delayed}
  {Teleoperation}},'' \emph{\BIBforeignlanguage{en}{IEEE Robotics and
  Automation Letters}}, vol.~5, no.~2, pp. 2981--2986, Apr. 2020. [Online].
  Available: \url{https://ieeexplore.ieee.org/document/8990033/}
\BIBentrySTDinterwordspacing

\bibitem{panganiban_transparency_2019}
\BIBentryALTinterwordspacing
A.~R. Panganiban, G.~Matthews, and M.~D. Long,
  ``\BIBforeignlanguage{en}{Transparency in {Autonomous} {Teammates}:
  {Intention} to {Support} as {Teaming} {Information}},''
  \emph{\BIBforeignlanguage{en}{Journal of Cognitive Engineering and Decision
  Making}}, p. 155534341988156, Nov. 2019. [Online]. Available:
  \url{http://journals.sagepub.com/doi/10.1177/1555343419881563}
\BIBentrySTDinterwordspacing

\bibitem{chen_increasing_2015}
\BIBentryALTinterwordspacing
T.~Chen, D.~Campbell, L.~F. Gonzalez, and G.~Coppin,
  ``\BIBforeignlanguage{en}{Increasing {Autonomy} {Transparency} through
  capability communication in multiple heterogeneous {UAV} management},'' in
  \emph{\BIBforeignlanguage{en}{2015 {IEEE}/{RSJ} {International} {Conference}
  on {Intelligent} {Robots} and {Systems} ({IROS})}}.\hskip 1em plus 0.5em
  minus 0.4em\relax Hamburg, Germany: IEEE, Sep. 2015, pp. 2434--2439.
  [Online]. Available: \url{http://ieeexplore.ieee.org/document/7353707/}
\BIBentrySTDinterwordspacing

\bibitem{lakhmani_exploring_2019}
\BIBentryALTinterwordspacing
S.~G. Lakhmani, J.~L. Wright, M.~R. Schwartz, and D.~Barber,
  ``\BIBforeignlanguage{en}{Exploring the {Effect} of {Communication}
  {Patterns} and {Transparency} on {Performance} in a {Human}-{Robot}
  {Team}},'' \emph{\BIBforeignlanguage{en}{Proceedings of the Human Factors and
  Ergonomics Society Annual Meeting}}, vol.~63, no.~1, pp. 160--164, Nov. 2019.
  [Online]. Available:
  \url{http://journals.sagepub.com/doi/10.1177/1071181319631054}
\BIBentrySTDinterwordspacing

\bibitem{matthews_individual_2019}
\BIBentryALTinterwordspacing
G.~Matthews, J.~Lin, A.~R. Panganiban, and M.~D. Long,
  ``\BIBforeignlanguage{en}{Individual {Differences} in {Trust} in {Autonomous}
  {Robots}: {Implications} for {Transparency}},''
  \emph{\BIBforeignlanguage{en}{IEEE Transactions on Human-Machine Systems}},
  pp. 1--11, 2019. [Online]. Available:
  \url{https://ieeexplore.ieee.org/document/8908731/}
\BIBentrySTDinterwordspacing

\bibitem{guznov_robot_2019}
\BIBentryALTinterwordspacing
S.~Guznov, J.~Lyons, M.~Pfahler, A.~Heironimus, M.~Woolley, J.~Friedman, and
  A.~Neimeier, ``\BIBforeignlanguage{en}{Robot {Transparency} and {Team}
  {Orientation} {Effects} on {Human}–{Robot} {Teaming}},''
  \emph{\BIBforeignlanguage{en}{International Journal of Human–Computer
  Interaction}}, pp. 1--11, Oct. 2019. [Online]. Available:
  \url{https://www.tandfonline.com/doi/full/10.1080/10447318.2019.1676519}
\BIBentrySTDinterwordspacing

\bibitem{daily_world_2003}
M.~Daily, Y.~Cho, K.~Martin, and D.~Payton, ``World embedded interfaces for
  human-robot interaction,'' in \emph{System {Sciences}, 2003. {Proceedings} of
  the 36th {Annual} {Hawaii} {International} {Conference} on}.\hskip 1em plus
  0.5em minus 0.4em\relax IEEE, 2003, pp. 6--pp.

\bibitem{wright_transparency_2015}
\BIBentryALTinterwordspacing
J.~L. Wright, ``\BIBforeignlanguage{en}{Transparency in {Human}-agent {Teaming}
  and its {Effect} on {Automation}-induced {Complacency}},''
  \emph{\BIBforeignlanguage{en}{Procedia Manufacturing}}, vol.~3, pp. 968--973,
  2015. [Online]. Available:
  \url{https://linkinghub.elsevier.com/retrieve/pii/S235197891500150X}
\BIBentrySTDinterwordspacing

\bibitem{wright2017agent}
J.~L. Wright, J.~Y. Chen, M.~J. Barnes, and P.~A. Hancock, ``Agent reasoning
  transparency: The influence of information level on automation induced
  complacency,'' US Army Research Laboratory Aberdeen Proving Ground United
  States, Tech. Rep., 2017.

\bibitem{wright_effects_2015}
\BIBentryALTinterwordspacing
J.~L. Wright, J.~Y. Chen, M.~J. Barnes, and M.~W. Boyce,
  ``\BIBforeignlanguage{en}{The {Effects} of {Information} {Level} on
  {Human}-{Agent} {Interaction} for {Route} {Planning}},''
  \emph{\BIBforeignlanguage{en}{Proceedings of the Human Factors and Ergonomics
  Society Annual Meeting}}, vol.~59, no.~1, pp. 811--815, Sep. 2015. [Online].
  Available: \url{http://journals.sagepub.com/doi/10.1177/1541931215591247}
\BIBentrySTDinterwordspacing

\bibitem{ghiringhelli_interactive_2014}
F.~Ghiringhelli, J.~Guzzi, G.~A. Di~Caro, V.~Caglioti, L.~M. Gambardella, and
  A.~Giusti, ``Interactive augmented reality for understanding and analyzing
  multi-robot systems,'' in \emph{Intelligent {Robots} and {Systems} ({IROS}
  2014), 2014 {IEEE}/{RSJ} {International} {Conference} on}.\hskip 1em plus
  0.5em minus 0.4em\relax IEEE, 2014, pp. 1195--1201.

\bibitem{mercado_intelligent_2016}
\BIBentryALTinterwordspacing
J.~E. Mercado, M.~A. Rupp, J.~Y.~C. Chen, M.~J. Barnes, D.~Barber, and
  K.~Procci, ``\BIBforeignlanguage{en}{Intelligent {Agent} {Transparency} in
  {Human}–{Agent} {Teaming} for {Multi}-{UxV} {Management}},''
  \emph{\BIBforeignlanguage{en}{Human Factors: The Journal of the Human Factors
  and Ergonomics Society}}, vol.~58, no.~3, pp. 401--415, May 2016. [Online].
  Available: \url{http://journals.sagepub.com/doi/10.1177/0018720815621206}
\BIBentrySTDinterwordspacing

\bibitem{jakobi1995noise}
N.~Jakobi, P.~Husbands, and I.~Harvey, ``Noise and the reality gap: The use of
  simulation in evolutionary robotics,'' in \emph{European Conference on
  Artificial Life}.\hskip 1em plus 0.5em minus 0.4em\relax Springer, 1995, pp.
  704--720.

\bibitem{vicon}
``{VICON} motion capture system,'' \url{http://vicon.com}, accessed:.

\bibitem{Pinciroli:SI2012}
C.~Pinciroli, V.~Trianni, R.~O'Grady, G.~Pini, A.~Brutschy, M.~Brambilla,
  N.~Mathews, E.~Ferrante, G.~{Di Caro}, F.~Ducatelle, M.~Birattari, L.~M.
  Gambardella, and M.~Dorigo, ``{ARGoS}: a modular, parallel, multi-engine
  simulator for multi-robot systems,'' \emph{Swarm Intelligence}, vol.~6,
  no.~4, pp. 271--295, 2012.

\bibitem{vuforia}
``{Vuforia} augmented reality,'' \url{http://vuforia.com}, accessed:.

\bibitem{engine2008unity}
U.~G. Engine, ``Unity game engine-official site,'' \emph{Online][Cited: October
  9, 2008.] http://unity3d. com}, pp. 1534--4320, 2008.

\bibitem{taylor2017situational}
R.~M. Taylor, ``Situational awareness rating technique (sart): The development
  of a tool for aircrew systems design,'' in \emph{Situational
  awareness}.\hskip 1em plus 0.5em minus 0.4em\relax Routledge, 1990, pp.
  111--128.

\bibitem{likert}
\BIBentryALTinterwordspacing
R.~LIKERT, ``A technique for the measurement of attitudes,'' \emph{Arch Psych},
  vol. 140, p.~55, 1932. [Online]. Available:
  \url{https://ci.nii.ac.jp/naid/10024177101/en/}
\BIBentrySTDinterwordspacing

\bibitem{hart1988development}
S.~G. Hart and L.~E. Staveland, ``Development of nasa-tlx (task load index):
  Results of empirical and theoretical research,'' in \emph{Advances in
  psychology}.\hskip 1em plus 0.5em minus 0.4em\relax Elsevier, 1988, vol.~52,
  pp. 139--183.

\bibitem{uggirala2004measurement}
A.~Uggirala, A.~K. Gramopadhye, B.~J. Melloy, and J.~E. Toler, ``Measurement of
  trust in complex and dynamic systems using a quantitative approach,''
  \emph{International Journal of Industrial Ergonomics}, vol.~34, no.~3, pp.
  175--186, 2004.

\bibitem{friedman1937use}
M.~Friedman, ``The use of ranks to avoid the assumption of normality implicit
  in the analysis of variance,'' \emph{Journal of the american statistical
  association}, vol.~32, no. 200, pp. 675--701, 1937.

\bibitem{black1976partial}
D.~Black, ``Partial justification of the borda count,'' \emph{Public Choice},
  pp. 1--15, 1976.

\end{thebibliography}

\end{document}